\documentclass[lettersize,journal]{IEEEtran}

\usepackage{custom}
\usepackage{amsmath,amsfonts}
\usepackage{algorithmic}

\usepackage{array}
\usepackage[caption=false,font=normalsize,labelfont=sf,textfont=sf]{subfig}
\usepackage{textcomp}
\usepackage{stfloats}
\usepackage{url}
\usepackage{verbatim}
\usepackage{graphicx}
\usepackage{cite}
\usepackage[normalem]{ulem}
\usepackage{mathabx}

\usepackage{comment}
\usepackage{xcolor}

\usepackage{siunitx}

\usepackage{nicematrix}
\usepackage{mathtools}
\usepackage{booktabs}
\usepackage{svg}
\usepackage{xifthen}
\usepackage{xparse}
\usepackage{bm}

\usepackage{tikz}
\tikzset{
  exterior size/.style args={#1x#2}{
    outer sep=auto,
    minimum width={#1-\pgflinewidth},
    minimum height={#2-\pgflinewidth}
  }
}

\definecolor{cvprblue}{rgb}{0.21,0.49,0.74}
\definecolor{green}{rgb}{0,0.6,0}
\usepackage[pagebackref,breaklinks,colorlinks,citecolor=cvprblue]{hyperref}

\newcommand\numberthis{\addtocounter{equation}{1}\tag{\theequation}}

\NewDocumentCommand{\amend}{m o}{%
  \textcolor{black}{#1}%
  \IfValueT{#2}{ \sout{#2}}%
}
\NewDocumentCommand{\amendDF}{m o}{%
  \textcolor{green}{#1}%
  \IfValueT{#2}{ \sout{#2}}%
}

\usepackage{algorithm2e}

\SetCommentSty{mycommfont}
\SetKw{Break}{break}
\SetKw{True}{True}
\SetKwComment{Comment}{\# }{}
\SetKwInOut{Inputs}{Inputs}\SetKwInOut{Outputs}{Outputs}

\newcommand\copyrighttext{%
  \footnotesize \textcopyright \the\year{} IEEE. Personal use of this material is permitted. Permission from IEEE must be obtained for all other uses, including reprinting/republishing this material for advertising or promotional purposes, collecting new collected works for resale or redistribution to servers or lists, or reuse of any copyrighted component of this work in other works.}

\newcommand\copyrightnotice{%
\begin{tikzpicture}[remember picture,overlay]
\node[anchor=south,yshift=10pt] at (current page.south) {\fbox{\parbox{\dimexpr0.75\textwidth-\fboxsep-\fboxrule\relax}{\copyrighttext}}};
\end{tikzpicture}%
}

\begin{document}
\newcommand{\stack}[2]{\left[#1_1, \ldots, #1_{#2} \right]}

\newcommand{\R}[1]{\mathbb{R}^{#1}}
\newcommand{\Rl}{\R{l}}
\newcommand{\Rkt}{\R{k \times 3}}
\newcommand{\rotgroup}{\mathrm{SO(3)}}
\newcommand{\motiongroup}{\mathrm{SE(3)}}

\newcommand{\set}[2][]{%
   \ifthenelse{ \isempty{#1} }                                 
      { \ensuremath{\left\{ #2 \right\} } }
      { \ensuremath{\left\{ #2 \right\}_{1 \leq i \leq #1} } }
}
\newcommand{\sX}{\mathcal{X}}
\newcommand{\sZ}{\bm{Z}}

\newcommand{\motion}[2][]{%
  \ifthenelse{ \isempty{#2} }                                 
      {
        \ifthenelse{ \isempty{#1} }                            
            { \ensuremath{\rho} }                                  
            { \ensuremath{\rho_{#1}} }                             
      }
      {
        \ifthenelse{ \isempty{#1} }                             
            { \ensuremath{\rho #2} }                
            { \ensuremath{\rho_{#1} #2} }          
      }
}

\newcommand\argmin[1]{\underset{#1}{\text{argmin}}\,}
\newcommand{\orbit}[1]{\mathcal{O}_{#1}}         
\newcommand{\norm}[1]{\left\lVert#1\right\rVert} 
\newcommand{\ra}[2]{\angle \left(#1, #2 \right) }

\newcommand{\alldots}[3]{#1 = #2,\ldots, #3}     

\newcommand{\lieK}{\bm{\mathrm{K}}}
\newcommand{\Rot}{\bm{\mathrm{R}}}
\newcommand{\rmF}{\mathrm{F}}
\newcommand{\Loss}{\mathrm{L}}

\newcommand{\Rho}{\bm{\rho}}
\newcommand{\bT}{\bm{T}}
\newcommand{\bD}{\bm{D}}
\newcommand{\bt}{\bm{t}}
\newcommand{\bX}{\bm{X}}
\newcommand{\bx}{\bm{x}}
\newcommand{\bP}{\bm{P}}
\newcommand{\bp}{\bm{p}}
\newcommand{\bQ}{\bm{Q}}
\newcommand{\bq}{\bm{q}}
\newcommand{\bz}{\bm{z}}
\newcommand{\bV}{\bm{V}}
\newcommand{\bk}{\bm{k}}
\newcommand{\bK}{\bm{K}}
\newcommand{\bRot}{\bm{\Rot}}
\newcommand{\bLoss}{\bm{\Loss}}

\newcommand{\dCD}{d_{\operatorname{BC}}}
\newcommand{\dSD}{d_{\operatorname{SC}}}
\newcommand{\dPSD}{d_{\operatorname{PSC}}}
\newcommand{\dNN}{d_{\operatorname{NN}}}

\newcommand{\Xs}{X^{\text{source}}}
\newcommand{\Xt}{X^{\text{target}}}

\newcommand{\LMSR}{\mathrm{FLAMES}}
\newcommand{\LossLMSR}{\bm{\Loss}^{\rmF}}
\newcommand{\lossLMSR}{\Loss^{\rmF}}
\newcommand{\RotLMSR}{\bm{\Rot}^{\rmF}}
\newcommand{\rotLMSR}{\Rot^{\rmF}}
\newcommand{\RhoLMSR}{\bm{\Rho}^{\rmF}}
\newcommand{\rhoLMSR}{\rho^{\rmF}}

\newcommand{\E}{\mathrm{e}}
\newcommand{\D}{\mathrm{d}}

\title{Multiview Point Cloud Registration via Optimization in an Autoencoder Latent Space}
\author{Luc Vedrenne, Sylvain Faisan, Denis Fortun%
\thanks{This work of the Interdisciplinary Thematic Institute HealthTech, as part of the ITI 2021-2028 program of the University of Strasbourg, CNRS and Inserm, was partially supported by IdEx Unistra (ANR-10-IDEX-0002) and SFRI (STRAT’US project, ANR-20-SFRI-0012) under the framework of the French Investments for the Future Program. It was also supported by the French National Research Agency (ANR) through the SP-Fluo project (ANR-20-CE45-0007).}%
\thanks{Luc Vedrenne (corresponding author: \href{mailto:vedrenne@unistra.fr}{vedrenne@unistra.fr}), Sylvain Faisan, and Denis Fortun are with the ICube Laboratory, IMAGeS team, UMR 7357, CNRS, University of Strasbourg, France.}
}

\IEEEpubid{978-1-5386-5541-2/25\$31.00~\copyright~2024 IEEE}

\maketitle

\copyrightnotice
\begin{abstract}
Point cloud rigid registration is a fundamental problem in 3D computer vision. In the multiview case, we aim to find a set of 6D poses to align a set of objects. Methods based on pairwise registration rely on a subsequent synchronization algorithm, which makes them poorly scalable with the number of views. Generative approaches overcome this limitation, but are based on Gaussian Mixture Models and use an Expectation-Maximization algorithm. Hence, they are not well suited to handle large transformations. Moreover, most existing methods cannot handle high levels of degradations. In this paper, we introduce POLAR (POint cloud LAtent Registration), a multiview registration method able to efficiently deal with a large number of views, while being robust to a high level of degradations and large initial angles. To achieve this, we transpose the registration problem into the latent space of a pretrained autoencoder, design a loss taking degradations into account, and develop an efficient multistart optimization strategy. Our proposed method significantly outperforms state-of-the-art approaches on synthetic and real data. POLAR is available at \href{https://github.com/pypolar/polar}{github.com/pypolar/polar} or as a standalone package which can be installed with \texttt{\small pip install polaregistration}.
\end{abstract}

\section{Introduction}
\label{sec:intro}

\IEEEPARstart{D}{ownstream} tasks in 3D computer vision often involve a rigid registration step \cite{Heydarian21, pcr_Blais95, Nuchter07, fusion_Newcombe15, kinectfusion_Newcombe15, 4dreconstruct_Tang21}, which consists of determining 6D rigid transformations to align objects. In the multiview context, the objective is to align a set of point clouds, rather than just a pair. This paper focuses on {\it object-level} registration, where each view represents the same object and the objective is to reconstruct an accurate 3D model of this reference  \cite{hybridmixture_Min20, Heydarian21}. In particular, our primary goal is to address the problem of  data acquired in a microscopy modality called SMLM (single molecule localization microscopy). The challenge of registration with this data is that the views have undergone a very high level of anisotropic noise and outliers, and a moderate level of occlusions. This is in contrast with {\it scene-level} applications, where the point clouds represent fragments of a large-scale scene obtained with LIDAR or RGB-D camera, with very low noise and outliers, but high occlusion levels. In what follows, we use the term {\it degradation} to refer to the alteration of the shape of a point cloud due to noise, occlusions or outliers.

\IEEEpubidadjcol
Multiview registration methods can be classified into two main categories. The first one, largely predominant, estimates all the pairwise relative motions, and then runs a subsequent synchronization algorithm to retrieve absolute poses from all the relative ones \cite{globmo_Govindu18, globmo_Torsello11, globmo_Maset17, globmo_Birdal18, globmo_Bernard15, globmo_Arrigoni16, globmo_Arrigoni14, globmo_ArieNachimson212, robmulti_Wang23}. This approach has three main limitations: (i) it poorly scales with the number of views, as it requires $\mathcal{O}(N^2)$ registrations to retrieve $N$ absolute poses, in addition to the cost of the synchronization; (ii) all failed pairwise registrations negatively impact the overall result; (iii) each pairwise registration is performed independently, without leveraging informations from other views. The second family of methods is the generative approach: a template of the reference object from which the views are observed is estimated, and all the views are registered onto this template \cite{Jian11, mlmd_Eckart15, em_Chui20, jrmpc_Evangelidis17, cpd_Myronenko10, filterreg_Gao19, scam_Zhu20, hybridmixture_Min20, tstudent_Ma22}. This is typically achieved by modeling the template as a probability distribution using a Gaussian Mixture Model (GMM), jointly estimated with the absolute poses using an Expectation-Maximization (EM) algorithm. These generative approaches have the primary benefit of simultaneously registering all point clouds, which mitigates the limitations of the synchronization approach.
However, they are only able to converge to local minima, which limits their applicability to refining poses of already coarsely aligned objects. Moreover, they cannot benefit from the robustness and flexibility of point cloud descriptors learned by neural networks. Hence, almost all state-of-the-art multiview registration methods are currently based on synchronization.

We propose to revitalize the generative approach by formulating the multiview registration problem entirely in the latent space of an autoencoder, which has been pretrained to reconstruct clean views from degraded ones. The interest is fourfold:
(i) as a generative approach, it enables the simultaneous registration of numerous views, (ii) the template is modeled by a global latent vector learned by a deep neural network, which arguably makes it more expressive than the usual mixture of Gaussians, (iii) leveraging a global descriptor enables correspondence-free registration, yielding more robustness to noise and occlusion than conventional methods based on feature matching, (iv) the latent space enables faster and more scalable optimization than its ambient counterpart.
Moreover, we design a loss that takes degradations into account, such that the estimated template is progressively restored during the optimization. Finally, we propose an optimization scheme able to retrieve the global optimum from potentially many local ones, which allows our method to handle arbitrarily large transformations between the views to register. 

We begin with a brief overview of existing registration methods related to our work and their limitations in \cref{sec:related_works}. In \cref{sec:method}, we describe our method, POLAR. In \cref{sec:motivation}, we provide a theoretical justification of the principle of POLAR. Implementation details are provided in \cref{sec:implem}. Extensive experiments across many challenging scenarios on both synthetic and real-world data are conducted in \cref{sec:experiments}, empirically demonstrating the benefits of our approach.
\section{Related works}
\label{sec:related_works}
We first introduce popular multiview registration methods, either based on synchronization or on a generative framework (\cref{sec:related_works_multiview}). We then focus on registration approaches that remain pairwise but nonetheless address specific challenges related to our work (\cref{sec:related_works_pairwise}).

\subsection{Multiview registration}\label{sec:related_works_multiview}

\noindent\textbf{Synchronization} The main challenge of motion synchronization is to limit the impact of pairwise registration failures. Hence, many synchronization algorithms seek to achieve robust synchronization, by relying on convex relaxation \cite{globmo_ArieNachimson212, globmo_Arrigoni14, globmo_Bernard15, globmo_Arrigoni16, globmo_Maset17}, Iterative Reweighted Least Squares \cite{globmo_Govindu18, robmulti_Wang23}, or by operating on the quaternion or $\mathfrak{se}(3)$ Lie algebras \cite{globmo_Torsello11, globmo_Birdal18, lieavg_Govindu2004}. Recently, some works have proposed to learn this synchronization \cite{synchrolearn_Huang19, deepmapping_Ding19} \amend{+ \cite{learn_irts_Yew21}}. In \cite{learnmulti_Gojcic20}, the whole process of pairwise registration and synchronization is learned end-to-end.

\noindent\textbf{Generative approach} Existing generative methods all rely on an EM algorithm, which is sensitive to initialization. The primary differences among these methods lie in how they model the reconstructed template onto which the objects are registered. Most generative methods rely on GMMs to represent the probability distribution of the reference model onto which the objects are registered \cite{Jian11, mlmd_Eckart15, em_Chui20, jrmpc_Evangelidis17, cpd_Myronenko10, filterreg_Gao19, scam_Zhu20}. Variants have been proposed to enhance robustness, by considering normal vector information with Hybrid mixture models \cite{hybridmixture_Min20}, by handling outliers with the Student's t-mixture model \cite{tstudent_Ma22} or Laplacian model \cite{robustMR_Zhang21}, or by performing fuzzy clustering \cite{fuzzyCG_Liao22, infoGM_Wang22}. 

\subsection{Pairwise registration}\label{sec:related_works_pairwise}

\noindent\textbf{Correspondence-based registration} The problem of registering a pair of objects is conventionally addressed through a three-step process, possibly iterated until convergence is achieved: (1) feature extraction, (2) feature matching, and (3) transformation estimation. In its simplest form, features correspond to point coordinates, matching is accomplished through a nearest neighbor projection, and the rigid transformation is estimated from correspondences by SVD. This is the ICP (Iterative Closest Point) algorithm \cite{icp_Besl92}, which has given rise to numerous variants \cite{icpvariant_Rusink01}.
Each of these three steps can be enhanced by deep learning algorithms. Numerous methods have sought to learn descriptive features, \amend{with an emphasis on rotation-invariant features}\cite{Ao2020SpinNetLA, Deng18, Yu2022RIGARA,  roitr_Yu22, Qin2022GeometricTF} \amend{or additional geometric properties}\cite{Bai2020D3FeatJL, Saleh2020GraphiteGF, perfectmatch_Gojcic19, evomultidesc_Wu24, egst_Yuan24}. \amend{The most recent ones leverage a geometric Transformer network}\cite{Yu2022RIGARA, Yew2022REGTREP, roitr_Yu22, egst_Yuan24}, \amend{often in a multi-scale coarse-to-fine approach} \cite{Qin2022GeometricTF, egst_Yuan24, mpct_Wu24}. Some further learn the matching procedure \cite{fcgf_Choy19, Deng2018PPFNetGC, Huang2020PREDATORRO, Li2021LepardLP, Qin2022GeometricTF, Saleh2022BendingGH, Yew2022REGTREP, Yu2021CoFiNetRC, 3dmatch_Zeng17, Zhang2023PCRCGPC, egst_Yuan24}, typically through \amend{a differentiable Sinkhorn algorithm}. Estimating the transformation from the established correspondences is typically done through the RANSAC algorithm \cite{ransac_Fischler81} and its variants \cite{Barth2017GraphCutR, Quan2020CompatibilityGuidedSC, Rusu2009FastPF, Yang2022CorrespondenceSW, Yang2021SACCOTSC, Yang2021TowardEA}.

\noindent\textbf{Correspondence-free registration} A distinct category of methods obviates the need for explicit feature matching by characterizing a point cloud with a single global descriptor, typically learned by a neural network \cite{pnlk_Aoki19, pcrnet_Sarode19, fmr_Huang2020, omnet_Xu21, Zhu2022CorrespondenceFreePC}. This global descriptor is anticipated to be more robust to noise, local shape variations or repetitive patterns. However, methods based on global descriptors have been restricted to local convergence and can only be applied after a first step of coarse alignment.

\noindent\textbf{Global convergence} To overcome the sensitivity to local minima, several works have developed global optimization strategies, such as Branch-and-Bound \cite{goicp_Yang16} and graduated non-convexity \cite{gradutednonconvexity_Yang16, fgr_Zhou16}, or designed richer descriptors to widen the basins of convergence \cite{Hartley07, Enqvist08, Olsson09}. 

\noindent\textbf{Registration in a latent space} Methods that register objects in a dedicated latent space can be traced back to pairwise correlation-based methods that operate in the GMM latent space: the source and target point clouds are both modeled as GMMs and the optimization is performed by minimizing a divergence (typically Kullback-Leibler) \cite{Jian11, mlmd_Eckart15, em_Chui20}. DeepGMR \cite{deepgmr_Yan20} extends this idea by letting neural networks learn the whole process. Some works proposed to use an autoencoder to learn relevant features in an unsupervised manner \cite{Xie20, Deng18} or even to register \cite{fmr_Huang2020, Shen22, lorax_Elbaz17, Zhu2022CorrespondenceFreePC}, but remain non-generative, and therefore pairwise.

\section{Method}
\label{sec:method}

\subsection{Notations}
In what follows, $\motion{\bX} = \left[\motion{\bx_1}, \ldots, \motion{\bx_k} \right] \in \Rkt$ denotes the element-wise application of a rigid motion $\motion{} \in \motiongroup$ to a point cloud $\bX = \left[\bx_1, \ldots, \bx_k\right] \in \Rkt$. The use of bold mathematical font indicates a stacking of previously defined variables, whereas calligraphic font is strictly reserved for sets whose elements are of varying sizes. For instance, $\bm{A}$ denotes a stacking $\left[ \bm{A}_1 \in \R{k}, \ldots, \bm{A}_n \in \R{k} \right] \in \R{n \times k}$ whereas $\mathcal{A}$ denotes a set $\left\{\bm{A}_1 \in \R{k_1}, \ldots, \bm{A}_n \in \R{k_n} \right\}$.

\subsection{Problem formulation}
\label{sec:problem}
Let us consider an unknown reference point cloud $\bX^\star$ from which $N$ views $\sX = \{\bX_1, \ldots, \bX_N \}$ are observed, oriented by rigid motions $\Rho^\star = \stack{\motion{}^\star}{N}$ and corrupted by degradations $\varphi_i$:
\begin{equation}
\label{eq:generative_registration_problem}
\bX_i = \varphi_i\left(\motion[i]{}^\star \bX^\star\right),
\end{equation}
for $\alldots{i}{1}{N}$. Generative multiview registration aims to jointly estimate the $N$ rigid motions $\Rho^\star$ and the reference point cloud $\bX^\star$. To make the problem feasible, the goal is to estimate a parametric model (typically a GMM in previous works) from which the reference $\bX^\star$ can be generated. Note that there exist several solutions to this problem, corresponding to all possible poses of $\bX^\star$. 

\subsection{\amend{Overview}}
\label{sec:overview}

\begin{figure}
    \centering
    \includegraphics[width=\linewidth]{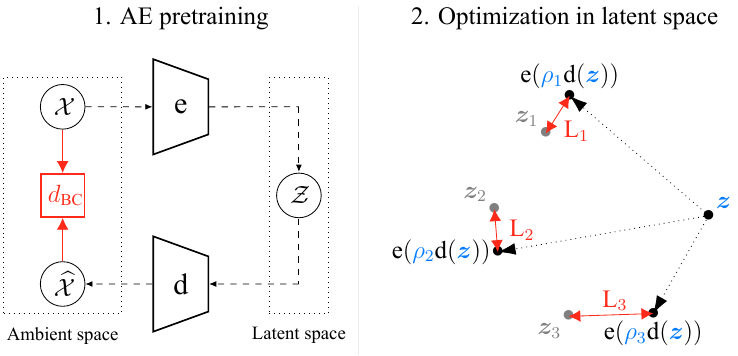}
    \caption{\amend{Overview of the proposed method. 1. Once and for all, before any registration, an autoencoder is trained to reconstruct point clouds (\cref{eq:bidirectional_chamfer}). 2. To register a set of views, an optimization problem is iteratively solved within the learnt latent space (\cref{eq:degraded_latent_optim_i}).}}
    \label{fig:overview_scheme}
\end{figure}

\amend{We propose a two-phase method. First, an autoencoder is trained to reconstruct point clouds, in order to learn a robust global descriptor (\cref{fig:overview_scheme}.1). This step is performed once and for all before registration, and the autoencoder does not need to be trained again for new point cloud data. The training is described in \cref{sec:AE_pretraining}. The actual registration is performed in the second phase, by optimizing a cost function defined in the latent space of the frozen autoencoder. We optimize this loss not only with respect to the pose parameters of the views, but also with respect to a latent vector that represents a reconstructed clean point cloud object, on which the views are registered. We describe our latent criterion in \cref{sec:latent_loss}. The optimization scheme to minimize this criterion is described in \cref{sec:optimization}.}

\subsection{Autoencoder pretraining} 
\label{sec:AE_pretraining}
 The fundamental component of POLAR is a pretrained autoencoder that provides a global descriptor of a point cloud in its latent space. Various architectures have been designed to represent a point cloud through a global descriptor \cite{pointnet_Qi2016, Xie20, Deng2018PPFNetGC}. As our approach is agnostic to the choice of the autoencoder, we consider in this section an abstract definition, and will present our specific implementation in \cref{sec:ae_training}. The general structure of an autoencoder is the composition of two differentiable functions, an encoder $\E: \Rkt \mapsto \Rl$ and a decoder $\D: \Rl \mapsto \Rkt$. We train these functions upstream to reconstruct and restore a point cloud $\bX$ degraded by a function $\varphi$, \ie to minimize $\dCD \left(\D \circ \E \circ \varphi(\bX), \bX\right)$ where $\dCD$ denotes the standard bidirectional Chamfer distance. For $\bP = \stack{\bp}{m} \in \R{m \times 3}$ and $\bQ = \stack{\bq}{n} \in \R{n \times 3}$, $\dCD$ is defined as
\begin{equation}\label{eq:bidirectional_chamfer}
    \dCD\left(\bP, \bQ\right) = \frac{1}{m} \sum_{i=1}^{m} \dNN(\bp_i, \bQ) + \frac{1}{n} \sum_{j=1}^{n} \dNN(\bq_j, \bP),
\end{equation}
where
\begin{equation}\label{eq:dnn}
    \dNN(\bp, \bQ) = \min_{\bq \in \bQ} \norm{ \bp - \bq}_2
\end{equation}
is the distance between $\bp$ and its nearest neighbor in $\bQ$. Once trained, this autoencoder is frozen, and the registration will be performed in its latent space. 

\subsection{\amend{Registration loss in the latent space}} 
\label{sec:latent_loss}
\begin{figure}
    \centering
    \includegraphics[width=\linewidth]{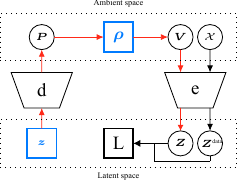}
    \caption{Graphical illustration of the loss computation in POLAR. $\sX = \left\{ \bX_1 \in \R{k_1}, \ldots, \bX_N \in \R{k_N}\right\}$ denotes the views to register. $\sZ^{\text{data}} = \left[e(\bX_1), \ldots, e(\bX_N) \right] \in \R{N \times l}$ is the matrix of their encodings. $\bP = \D(\bz) \in \Rkt$ is the estimated template. $\bV = \left[\motion[1]{\bP}, \ldots, \motion[N]{\bP} \right] \in \R{N \times k \times 3}$ denotes the views obtained by applying the estimated motions $\Rho$ to the estimated template $\bP$. Finally, $\sZ = \left[\E(\motion[1]{\bP}), \ldots, \E(\motion[N]{\bP}) \right] \in \R{N \times l}$ is the matrix of the latent vectors obtained by encoding these estimated views.}
    \label{fig:polar_scheme}
\end{figure}

\paragraph{Clean data}
We first consider the case of data without degradation. It follows from \cref{eq:generative_registration_problem} that the registration task reduces to finding $\bX^\star$ and $\Rho^\star$ such that $\bX_i = \motion[i]{}^\star \bX^\star$. Instead of directly comparing the point clouds $\bX_i$ and $\motion[i]{}^\star \bX^\star$, we compare their latent representation. We seek $\bX^\star$ and $\Rho^\star$ such that $\E(\bX_i) = \E(\motion[i]{}^\star \bX^\star)$. Furthermore, the template itself is expressed through its latent representation: we estimate a latent vector $\bz$ such that after decoding, $\D(\bz)$ represents $\bX^\star$. Thus, the encoding of a view to register $\E(\bX_i)$ is compared with the encoding of the reconstructed template $\D(\bz)$, rigidly transformed with the estimated pose $\motion[i]{}$, which writes $\E(\motion[i]{\D(\bz)})$. Hence, the optimization problem to solve is
\begin{equation}
\hat{\bz}, \hat{\Rho} = \argmin{\bz, \Rho} \Loss\left(\bz, \Rho \right)
\end{equation}
with
\begin{equation}\label{eq:latent_optim}
   \Loss\left(\bz, \Rho \right) = \sum_i^N \big\lVert \E\left(\motion[i]{\D(\bz)}\right) - \E(\bX_i) \big\rVert_2^2.
\end{equation}
A graphical model of the computation of this loss is presented in \cref{fig:polar_scheme}.

POLAR can be viewed as the combination of generative and deep correspondence-free approaches. It is generative because it estimates a parameterized representation of the object $X^\star$, which brings the benefits of simultaneous registration of all the point clouds. Since this parameterization is a global descriptor, we also avoid the need for local correspondences. Furthermore, this global descriptor is obtained from an autoencoder and thus exhibits an increased robustness and modeling expressiveness, in particular compared to the standard GMM representation. 

\amend{The loss of \cref{eq:latent_optim} is designed for registration of non-degraded point clouds. In the following paragraphs, we extend this loss to handle anisotropic noise, partial visibility, and outliers.}

\begin{figure}
    \centering
    \subfloat[\label{fig:mask_crop}]{%
       \includegraphics[width=0.49\linewidth]{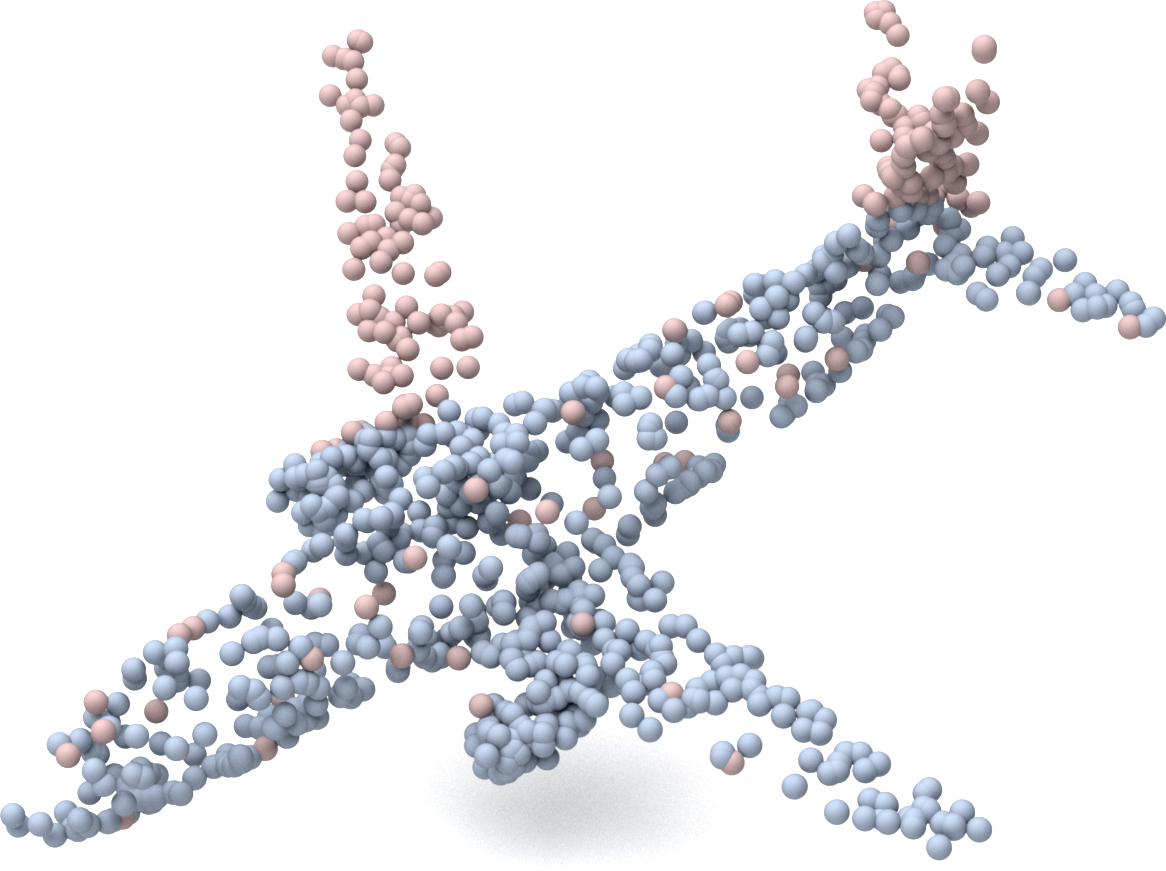}}
    \hfill
    \subfloat[\label{fig:mask_outliers}]{%
        \includegraphics[width=0.49\linewidth]{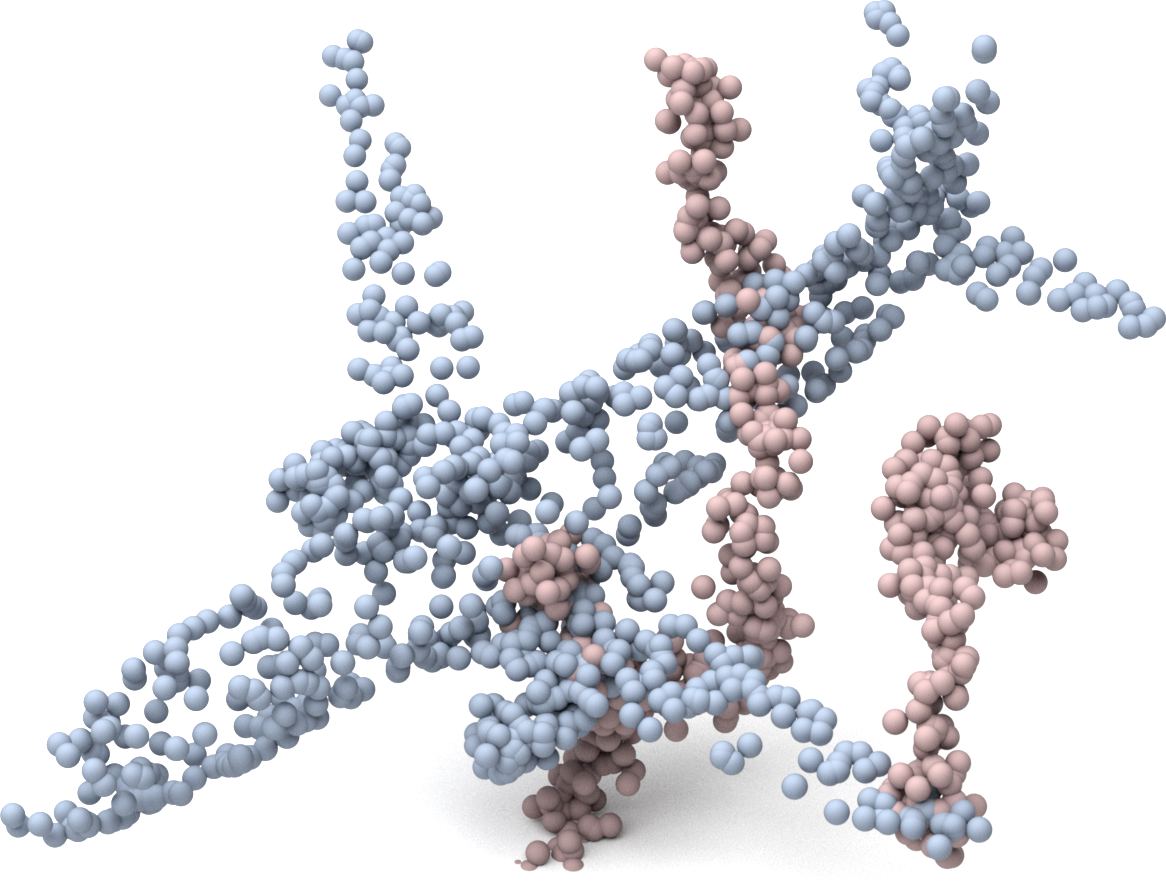}}
    \caption{Visual results of the masking operations (\cref{sec:latent_loss}.c, \cref{sec:latent_loss}.d) in case of occlusion and outliers. Red points in the template \textbf{(a)} and in the view \textbf{(b)} will be discarded for the loss computation.}
    \label{fig:real_masks}
\end{figure}

\paragraph{Anisotropic Noise} We first consider the case of noisy data. We denote $\nu_i$ the function that randomly noises each point of a cloud, under the \iid assumption, such that our degradation model is $\varphi_i(\bX) = \nu_i(\bX)$. We make no hypothesis regarding the nature of this noise. In particular, it can be anisotropic and different for each view. Using the loss \eqref{eq:latent_optim} in this case would lead to a noisy reconstruction $\D(\bz)$ that would fit the noisy data $\bX_i$. Therefore, in order to enforce the denoising of $\D(\bz)$, we follow the generative model in \cref{eq:generative_registration_problem} and account for this noise by applying $\nu_i$ to the reconstructed point cloud to create a noisy reconstruction. Thus, the loss \eqref{eq:latent_optim} becomes
\begin{equation}\label{eq:latent_optim_noisy}
    \Loss\left(\bz, \Rho \right) = \sum_i^N \big\lVert \E\big(\nu_i(\motion[i]{\D(\bz))}\big) - \E(\bX_i) \big\rVert_2^2.
\end{equation}
Note that this loss is not deterministic. However, since one realisation of noise is added to each point, when a cloud is composed of a fairly high amount of points, the variability at the shape scale remains low enough for the optimization to converge. 
\paragraph{Partial visibility} We now consider the case where the views are partially occluded: some points of each view are masked out. An illustration of this degradation can be seen in \cref{fig:mask_crop}, where red points are occluded. Since the template is estimated from many views, each occluded in a unique way, it should be possible to reconstruct a complete object. Therefore, some regions of the complete reconstructed object $\motion[i]{\D(\bz)}$ will not have correspondences in the view $\bX_i$, which introduces undesirable discrepancies in their encoding $\E(\motion[i]{\D(\bz)})$ and $\E(\bX_i)$ and affects the computation of the loss \eqref{eq:latent_optim}. To prevent this, we estimate the parts of $\motion[i]{\D(\bz)}$ missing in $\bX_i$, in order to mask them in $\motion[i]{\D(\bz)}$. We denote this masked template as $\mathbf{M}^v_{\bX_i}\left(\motion[i]{\D(\bz)}\right)$, where $v \in [0, 1]$ is a parameter representing the proportion of occluded points in the views. The loss \eqref{eq:latent_optim} thus becomes
\begin{equation}\label{eq:latent_optim_crop}
\Loss\left(\bz, \Rho \right) = \sum_i^N \big\lVert \E\left( \mathbf{M}^v_{\bX_i}\left(\motion[i]{\D(\bz)}\right) \right) - \E(\bX_i) \big\rVert_2^2.
\end{equation}

\begin{figure}
    \centering
    \includegraphics[width=\linewidth]{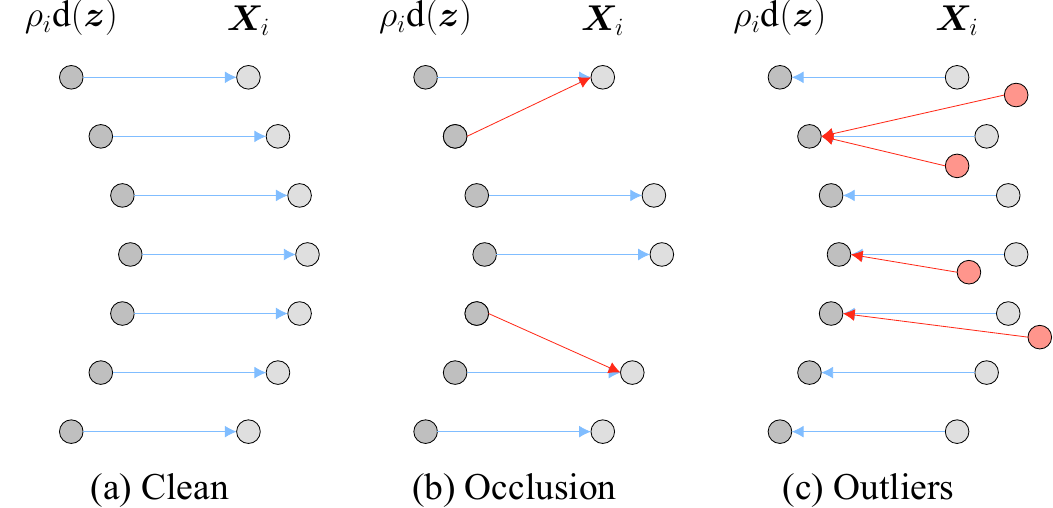}
    \caption{Schematic representation of nearest neighbor distances in three scenarios. The estimated template in the $i$-th pose $\motion[i]{\D(\bz)}$ is translated away from the corresponding view $\bX_i$ for visualization purpose but should be seen as superimposed, such that blue arrows denote null distances. \textbf{(a)} When the two objects are identical, all distances are null. \textbf{(b)} In case of occlusion, the distance from a point in the template to its nearest neighbor in the view is non-zero if and only if this point is occluded in the view (red arrows). \textbf{(c)} Similarly, the distance from a point in the view to its nearest neighbor in the template is non-zero if and only if this point is not part of the template, \ie if and only if it is an outlier (red arrows).}
    \label{fig:masks}
\end{figure}

To compute the mask $\mathbf{M}^v_{X_i}\left(\motion[i]{\D(\bz)}\right)$, we leverage the vector of nearest neighbor distances
\begin{equation}
\bD_{\bP \rightarrow \bQ} = \left[ \dNN(\bp_1, \bQ), \ldots, \dNN(\bp_m, \bQ) \right]
\end{equation}
where $\dNN(\bp_k, \bQ)$ is defined in \cref{eq:dnn}. The elements of the nearest neighbor distances $\bD_{\motion[i]{\D(\bz)} \rightarrow \bX_i}$ from the complete reconstructed template $\motion[i]{\D(\bz)}$ to the observed occluded point cloud $\bX_i$ are illustrated in \cref{fig:masks}.b. In the absence of noise, and assuming a perfect estimation of the pose, $\bD_{\motion[i]{\D(\bz)} \rightarrow \bX_i}$ is zero for visible points of $\bX_i$ and non zero for occluded ones. Hence, $\bD_{\motion[i]{\D(\bz)} \rightarrow \bX_i}$ can be used to select the points to mask: for a view $\bX_i$ and assuming that the proportion of occluded points $v$ is known, the proportion $v$ of elements of $\bD_{\motion[i]{\D(\bz)} \rightarrow \bX_i}$ with the highest values are considered as occluded points and are masked.

\paragraph{Outliers} Finally, we now describe how we handle the presence of outliers in the views. In \cref{fig:mask_outliers} we show an example of outliers on a view. In that case, we follow the same reasoning as for occluded data. As we seek a template $\motion[i]{\D(\bz)}$ without outliers, the outliers of the views $\bX_i$ have no correspondence in the template, and this will be reflected in their encodings. Thus, we estimate parts of $\bX_i$ missing in $\motion[i]{\D(\bz)}$ and mask them before computing the loss. Let $\mathbf{M}^o_{\motion[i]{\D(\bz)}}\left(\bX_i\right)$ denotes the masked view, where $o \in [0, 1]$ is a parameter representing the proportion of outliers in each view. The loss with outliers handling writes
\begin{equation}
    \Loss\left(\bz, \Rho \right) = \sum_i^N \big\lVert \E(\motion[i]{\D(\bz)}) - \E\big( \mathbf{M}^o_{\motion[i]{\D(\bz)}}\left(\bX_i\right) \big) \big\rVert_2^2.  
\end{equation}
As for occluded data, we compute the mask $\mathbf{M}^o_{\motion[i]{\D(\bz)}}\left(\bX_i\right)$ using the vector of nearest neighbor distances. But here we compute the distance from the view to the template $\bD_{\bX_i \rightarrow \motion[i]{\D(\bz)}}$. \cref{fig:masks}.c illustrates the values of this vector without noise and assuming perfect pose estimation: $\bD_{\bX_i \rightarrow \motion[i]{\D(\bz)}}$ is zero for inliers and non zero for outliers. Hence, assuming that the outliers ratio $o$ is known, a point $\bx$ in $\bX_i$ is masked if its distance $\dNN(\bx, \motion[i]{\D(\bz)})$ is above the $1 - o$ percentile of $\bD_{\bX_i \rightarrow \motion[i]{\D(\bz)}}$. 
\paragraph{Combining degradations} In the previous sections, we have presented how the loss \eqref{eq:latent_optim} can be extended to take into account three types of degradation separately. However, in real scenarios, data to register can present any combination of these degradations. If the views $\bX_i$ are noised with a function $\nu_i$, a ratio $1 - v$ of each view is occluded, and there is a ratio $o$ of outliers points in each view, we use the loss

\begin{equation}\label{eq:degraded_latent_optim}
    \Loss\left(\bz, \Rho \right) = \sum_i^N \Loss_i\left(\bz, \Rho \right)
\end{equation}
with
\begin{equation}\label{eq:degraded_latent_optim_i}
\Loss_i\left(\bz, \Rho \right) = 
\Big\lVert \E\Big( \mathbf{M}^v_{X_i}\big(\nu_i(\motion[i]{\D(\bz)})\big) \Big) - \E\big(\mathbf{M}^o_{\nu_i(\motion[i]{\D(\bz)})}(X_i) \big) \Big\rVert_2^2.
\end{equation}
This approach allows us to incorporate prior knowledge of the degradation model into the criterion.

\paragraph{Regularization}
\label{sec:regularization}
We empirically found that when dealing with highly degraded data, POLAR converges towards a template in which the distribution of points across the surface is not uniform: there are regions of high point density and regions of low point density. To penalize this behaviour, we add a regularization term that corresponds to the standard deviation of local point density. Precisely, given a point cloud $\bX \in \Rkt$, let $b_r(\bx, \bX)$ be the number of points in $\bX$ lying inside the ball of center $\bx$ and radius $r$ (\ie the number of neighbors of $\bx$). The mean point density of the point cloud $\bX$ is
\begin{equation}
    \bar{b}_r(\bX) = \frac{1}{k}\sum_{\bx \in \bX}b_r(\bx, \bX)
\end{equation}
and its variance is
\begin{equation}
    \sigma_{b_r}^2(\bX) = \frac{1}{k - 1}\sum_{\bx \in \bX} \left(b_r(\bx, \bX) - \bar{b}_r(\bX)\right)^2.
\end{equation}
We define the regularization term $R(\bz)$ as
\begin{equation}
    R(\bz) = \sigma_{b_r}(\D(\bz))
\end{equation}
and the final loss optimized in POLAR is 
\begin{equation}\label{eq:final_latent_optim}
    \Loss^f(\bz, \Rho) = \Loss(\bz, \Rho) + \lambda R(\bz)
\end{equation}
where $\lambda$ is a weighting hyperparameter.

\subsection{Optimization procedure}\label{sec:optimization}
\begin{figure}
\centering
\includegraphics[width=\linewidth]{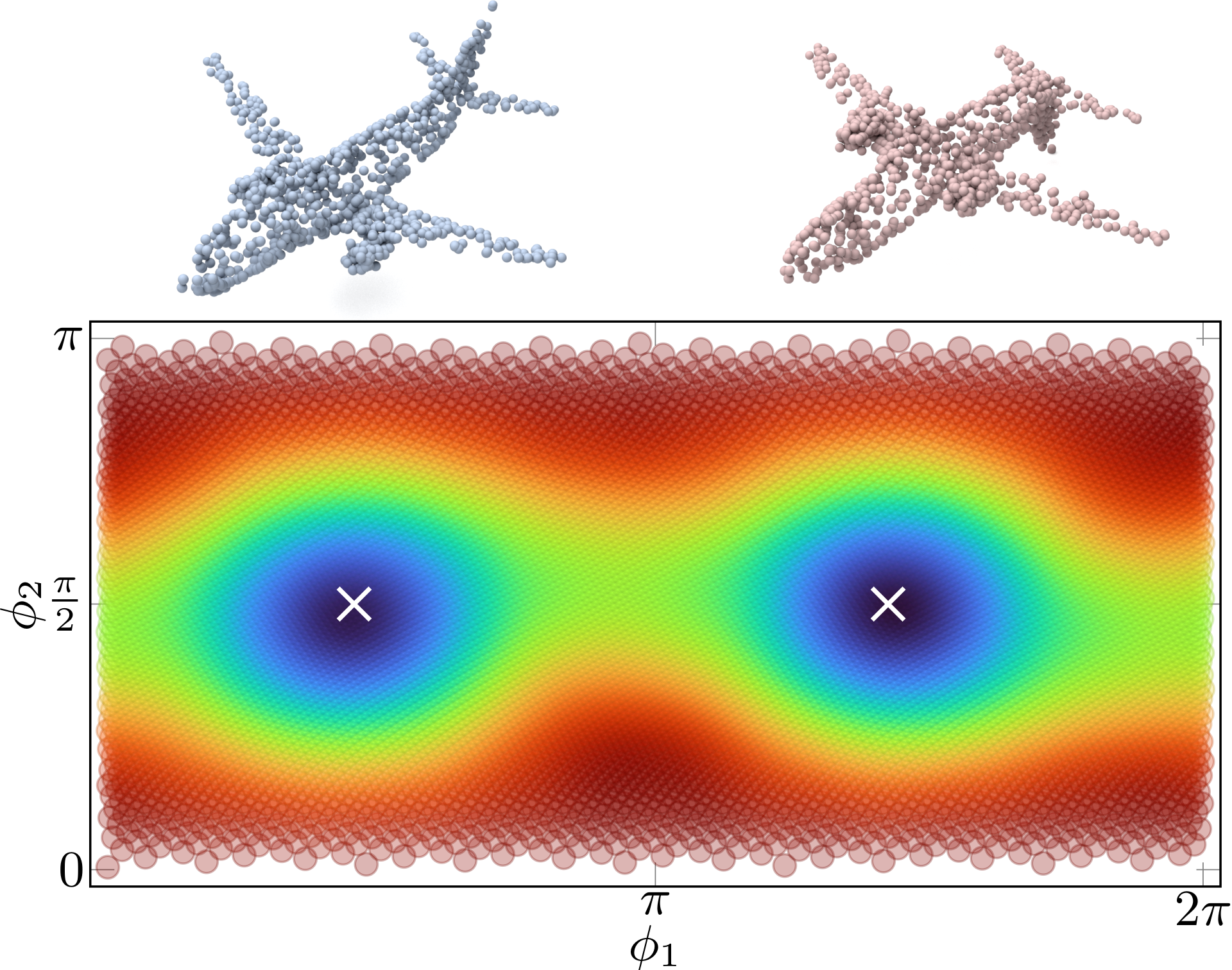}
\caption{Landscape of the loss in \cref{eq:latent_optim} with respect to the two first Euler angles (the loss is summed over the third Euler angle). The white crosses are the results of the $\rotgroup$ local minima search (\cref{sec:optimization}). In the case of an airplane, two minima coexist, corresponding to a $\ang{180}$ rotation along the fuselage axis.}
\label{fig:airplane_energy}
\end{figure}

We now describe how we minimize the criterion $\Loss^f$ in \cref{eq:final_latent_optim}. This optimization procedure is a hard task due to the highly non-convex nature of the problem. This is a fundamental issue in registration algorithms: as real-world objects often have symmetries or near-symmetries, discrepancy measures exhibit several local minima. This is illustrated in \cref{fig:airplane_energy} for an airplane, where 2 minima exist, corresponding to a $\ang{180}$ rotation along its fuselage axis (\cref{fig:airplane_energy}). As gradient-based optimization may be trapped in local minima, we design an optimization procedure able to retrieve the global minimum from potentially many local ones. Our approach employs a multistart strategy, detailed in the following subsections. In summary, we perform multiple gradient descents in parallel, starting from various plausible initializations, aiming to span all basins of attraction. These plausible starts are determined by a coarse exhaustive search described in \cref{sec:optimization}.a. The whole procedure is repeated until convergence.
\paragraph{Finding LocAl Minima ovEr $\rotgroup$ ($\LMSR)$} \label{sec:flames} Before decribing our full optimization pipeline, we present a novel necessary tool for finding local minima of the loss with respect to the rotations only, that we name $\LMSR$. These local minima are found by an exhaustive search over a uniform sampling of $L$ rotations in $\rotgroup$ denoted $\rotgroup_L$. Note that obtaining such a sampling is a hard problem, which we tackle using the Super Fibonacci spirals \cite{fibosampling_Alexa22}. A $k$-neighborhood graph is then defined in $\rotgroup_L$, using the relative angle, a geodesic distance on the manifold of rotations:
\begin{align*}\label{eq:relative_angle}
    \angle : \rotgroup \times \rotgroup &\mapsto [0, \pi]\\
    \Rot_1, \Rot_2 &\mapsto \arccos{\left(\frac{\mathrm{tr}(\Rot_1 \Rot_2^\top) - 1}{2}\right)}. \numberthis
\end{align*}
This \emph{k-nn} graph only depends on $L$ and $k$ and as such, can be pre-computed once and for all. We fix the current estimation of the template $\bz$. Thus, the terms $\Loss_i\left(\bz, \Rho\right)$ \cref{eq:degraded_latent_optim_i} of the criterion \eqref{eq:degraded_latent_optim} become independent. For each view, we fix the translation $t_i$ as well, and compute the criterion $\Loss_i\left(\bz, \Rho\right)$ \eqref{eq:degraded_latent_optim_i} for each sampled rotation. If a rotation has the lowest cost in its $k$-neighborhood, it is a local minimum for this view.
An example of loss landscape and local minima obtained with this procedure in the case of an airplane is illustrated in \cref{fig:airplane_energy}.

\paragraph{Initialization} Given a matrix $\sZ^{\text{data}} = \stack{\bz}{N}$ of $N$ latent vectors of views to register, the parameters to estimate $\bz$ and $\Rho$ are initialized as follows. The template $\bz_{\text{init}}$ is set to the medoid of the latent vectors $\sZ^{\text{data}}$, \ie the latent vector whose sum of dissimilarities to all the others is minimal:
\begin{equation}\label{eq:medoid}
    \bz_{\text{init}} = \argmin{\bz \in \sZ^{\text{data}}} \sum_{i=1}^N \lVert \bz - \bz_i \rVert_2
\end{equation}
The translations $\bt$ are set to zeros since the views are centered beforehand (\cf \cref{sec:dnormalization}). The rotations are initialized using the $\LMSR$ procedure described above, taking the local minimum of minimal cost for each view.
\paragraph{Joint gradient descent} The parameters $\bz$ and $\Rho$ are jointly updated to minimize the criterion \eqref{eq:final_latent_optim} by gradient descent, using the Adam algorithm \cite{adam_kingma14} with decoupled weight decay (AdamW) \cite{adamW_Loshchilov19}. The implementation is based on PyTorch's automatic differentiation engine \cite{pytorch_paszke19}.
\paragraph{Parallel multistart} 
Using the current estimation of $\bz$ and $\bt$, the $\LMSR$ procedure is used to find the $M$ best local minima over the rotation sampling $\rotgroup_L$ for each of the $N$ views. This results in $N \cdot M$ initializations, from which as many gradient descents are executed in parallel. In this step, we only optimize the rigid motions $\Rho$ and keep the current estimated template $\bz$ fixed. Note that this amounts to no more than $N \cdot M \cdot 6$ parameters. Even when registering a large number of views, this remains a relatively low-dimensional problem, which allows us to run all the gradient descents in parallel. After convergence, $M$ new poses and losses are obtained for each of the $N$ views. First, we select the new pose of minimal loss for each view. We note $\RhoLMSR = \stack{\rhoLMSR}{N}$ the obtained new poses and $\LossLMSR = \stack{\lossLMSR}{N}$ their corresponding losses. Given the current poses $\Rho = \stack{\motion[]{}}{N}$ and their corresponding losses $\bLoss = \stack{\Loss}{N}$, a new pose replaces the current one if it has a lower loss. Formally, the update function $u$ writes
\begin{equation}\label{eq:loss_improv}
    u(\rhoLMSR_i, \motion[i]{}) = 
    \begin{cases}
        \rhoLMSR_i   & \text{if}\quad \lossLMSR_i \leq \Loss_i \\
        \motion[i]{} & \text{if}\quad \lossLMSR_i > \Loss_i. \\
    \end{cases}
\end{equation}

\paragraph{Full procedure} After the first $\LMSR$ and parameters initialization, the sequence (i) joint gradient descent, (ii) $\LMSR$, (iii) parallel multistart is repeated until a stopping criterion is met. To design this criterion, we detect if we have escaped from a local minimum after the multistart step, right before the update \eqref{eq:loss_improv}. This is achieved by verifying if a new rotation is more than $T_{\Rot}$ away from the current one and if its loss is better than the current one. Let $\RotLMSR = \stack{\rotLMSR}{N}$ and $\bRot = \stack{\Rot}{N}$ denote the rotation part of the new poses $\RhoLMSR$ and of the current ones $\Rho$ respectively. The function $s$ detecting an escape from a local minimum is defined as
\begin{equation}\label{eq:rot_novelty}
    s(\rhoLMSR_i, \rho_i) = 
    \begin{cases}
        1 & \text{if}\; \lossLMSR_i \leq \Loss_i \;\text{and}\; \ra{\rotLMSR_i}{\Rot_i} \geq T_{\Rot}  \\
        0 & \text{otherwise}. \\
    \end{cases}
\end{equation}
The full optimization procedure is executed until no escape is detected, \ie until
\begin{equation}
    \sum_{i=1}^N s(\rhoLMSR_i, \rho_i)  = 0.
\end{equation}
The POLAR optimization scheme is summarized in \cref{alg:polar_optim}.  
\begin{algorithm}
\DontPrintSemicolon
\SetNoFillComment
\KwData{$\sZ^{\text{data}} = \E(\sX),\, \rotgroup_L$}
\KwResult{$\hat{z} ,\hat{\bRot}, \hat{\bt}$}
$z \gets \operatorname{medoid}(\sZ^{\text{data}})$\;
$\bt \gets \bm{0}$\;
$\bRot \gets \LMSR_{\text{top}\,1}(z, \bt, \sZ^{\text{data}}, \rotgroup_L)$\;
\While{not converged}{
$z, \bRot, \bt, \bLoss \gets \operatorname{joint\,gradient\,descent}(z, \bRot, \bt, \sZ^{\text{data}})$\;
$\RotLMSR \gets \LMSR_{\text{top}\,M}(z, \bt, \sZ^{\text{data}}, \rotgroup_L)$\;
$\RhoLMSR, \LossLMSR \gets \operatorname{multistart}(z, \RotLMSR, \bt, \sZ^{\text{data}}, \bLoss)$\;
$n \gets \sum_{i=1}^N s(\RhoLMSR_i, \Rho_i)$ \eqref{eq:rot_novelty}\; 
$\Rho \gets u(\RhoLMSR, \Rho)$ \eqref{eq:loss_improv}\;
\If{$n = 0$}{
\textit{converged}\;
}
}
$\hat{z}, \hat{\bRot}, \hat{\bt} \gets \operatorname{joint\,gradient\,descent}(z, \bRot, \bt, \sZ^{\text{data}})$\;
\caption{POLAR optimization algorithm. The notation $\LMSR_{\text{top}\,k}$ means that the $k$ best local minima for each view are selected following the FLAMES procedure of \cref{sec:optimization}.}
\label{alg:polar_optim}
\end{algorithm}

\section{Motivation}
\label{sec:motivation}
In this section, we interpret and motivate our model in \eqref{eq:latent_optim} from a differential geometry perspective.

The set of rigid motion $\motiongroup$ is endowed with a structure of Lie group. This property involves that the orbit of a point cloud $\bX$, denoted as $\orbit{\bX} = \left\{ \motion[]{\bX}\,,\, \rho \in SE(3) \right\}$, is a compact and smooth manifold of dimension at most 6 \cite{manifolds_Lee12}. In the absence of degradation, the views to register would be a sampling of the unknown template's orbit $\orbit{\bX^\star}$. In practice, data corruption moves away the observed views $\bX_i$ from the orbit $\orbit{\bX^\star}$. Our loss \eqref{eq:latent_optim} is defined in the autoencoder's latent space and it aims at estimating $\E(\orbit{\bX^\star})$, where $\E\left(\orbit{\bX}\right) = \left\{ \E(\motion[]{X})\,,\, \motion[]{} \in \motiongroup \right\}$ is the encoding of the orbit in the latent space. One interest of performing the registration in the latent space is that the encoder can be guided to produce a latent representation that is robust to data corruption. Of course, the encoder is not strictly invariant to degradation. Nevertheless, it is still robust to a certain extent, such that $\E(\varphi(\bX))$ should be closer to $\E(\bX)$ than $\varphi(\bX)$ is to $\bX$. 

In order to enable efficient gradient-based optimization in the latent space, it is crucial that $\E(\orbit{\bX})$ preserves the manifold structure of $\orbit{\bX}$. Since the image of a manifold under an embedding remains a manifold, it is sufficient for $\E$ to be an embedding (an injective function whose differential is injective). However, its smoothness is enough to ensure that $\E(\orbit{\bX})$ is "almost" a smooth manifold thanks to the following result \cite{topodiff_Hirsh76, manifolds_Lee12}:

\medskip
\noindent\textbf{Corollary of the Whitney embedding theorem} Suppose $\Gamma$ is a compact smooth n-manifold with or without boundary. If $l \geq 2n + 1$, then every smooth map from $\Gamma$ to $\Rl$ can be uniformly approximated by embeddings.
\medskip

\noindent With $\Gamma=\orbit{\bX}$, this corollary asserts that the encoder $\E$ can be uniformly approximated by an embedding (for any $\varepsilon > 0$, there exists an embedding $\mathrm{g}$ such that  $\lVert \mathrm{g} - \E \rVert_\infty < \varepsilon$), provided that $\E$ is a smooth function and the latent space has dimension $l > 13$. Thus, the set $\E\left(\orbit{\bX}\right)$ is arbitrarily close to a manifold. %
\section{Implementation details}
\label{sec:implem}

\subsection{Autoencoder architecture and training}
\label{sec:ae_training}
\paragraph{Architecture}
We define our encoder as a simple PointNet architecture \cite{pointnet_Qi2016} where we removed the T-net module. Our decoder is an MLP that takes the global feature of PointNet as input and outputs a point cloud. Batch norm is used for all layers with ReLU. DropOut layers are used for the decoder. Note that while PointNet is theoretically able to process point clouds of varying sizes, it is not trivial to come up with an implementation that actually allows it. A naive way would be to duplicate some points to pad each point cloud so that the autoencoder receives inputs of a fixed size. Unfortunately, the invariance to point duplication only holds true in the absence of Batch Normalization layers. To fully allow varying sizes without computational overhead while maintaining the use of Batch Normalization layers, we employ the message passing scheme from the \texttt{\small torch geometric} library \cite{FeyPyG2019} for our implementation. Note that any other global autoencoder could be integrated to POLAR.
\paragraph{Training}
We train a single autoencoder once and for all, and use it in all subsequent experiments, whether they involve simulated or real data, and regardless of the degradations. 
Our autoencoder is trained on the full ModelNet40 \cite{shapenet_Song15} training set. The training is performed for 200 epochs with the AdamW \cite{adamW_Loshchilov19} optimizer. The initial learning rate is $1e^{-3}$. The learning rate is divided by a factor of $2$ when the loss does not improve for $10$ consecutive epochs. We then process the data as follows. First, basic pre-processing steps are applied: $1024$ points are randomly sub-sampled (ModelNet is originally composed of dense point clouds of $5000$ points), centered, and normalized to lie exactly in the unit sphere. Then, a random pose is applied. To obtain an autoencoder able to reconstruct objects in arbitrary poses, it is crucial to uniformly sample $\rotgroup$, without relying on a discretization. To achieve this, we do not sample uniformly each Euler's angle, as it does not result in a uniform sampling over $SO(3)$. Instead, we leverage the exponential map from the Lie algebra to the underlying group $\rotgroup$. We sample an axis $\bk = \left[k_x, k_y, k_z \right]$ on the unit sphere and an angle $\theta \sim \mathcal{U}(0, \pi)$. The corresponding element $\lieK$ of the Lie algebra is
\begin{equation}
\lieK = 
\begin{bmatrix}
0 & -k_z & k_y \\
k_z & 0 & -k_x \\
- k_y & k_x & 0 \\
\end{bmatrix}
\end{equation}
and its associated rotation $\Rot$ in $\rotgroup$ is
\begin{equation}
    \Rot = \bm{I}_3 + \sin{\theta}\,\lieK + \left(1 - \cos{\theta}\right)\lieK^2.
\end{equation}
Finally, degradations are applied with the following policy:
\begin{enumerate}
    \item \textbf{Jit}: Add a centered multivariate Gaussian.
    \item \textbf{Plane Cut}: Sample a plane normal, and retain points close enough to this plane, so that a visibility ratio $v$ of the object is retained, with $v \sim \mathcal{U}(0.7, 1)$. \cite{rpmnet_Yew20, dcp_Wang19, Qin2022GeometricTF}
    \item \textbf{Center} \& \textbf{Normalize}: Center and scale to lie exactly within the unit sphere.
\end{enumerate}
On the standard ModelNet40 training set, the training takes about 20mn on a single NVIDIA GeForce RTX 3090. 

\subsection{Data normalization for registration}
\label{sec:dnormalization}
While it is common to normalize point clouds by centering and rescaling them to fit within a unit sphere, it is crucial for rigid registration to apply the same scaling factor for each view to be registered in order to ensure that point clouds maintain their relative sizes. This is achieved by independently calculating the scaling factor for each view and then normalizing them using the smallest previously calculated scaling factor.
\subsection{Hyperparameter setting}
For the \textit{k-nn}  graph over $\rotgroup$, we sample $L = 5e^4$ rotations and use $k=256$ neighbors. The multistart step considers the top $M=8$ minima per view, and $T_{\Rot}$ is set to $15^{\circ}$ to detect escapes from local minima. The decoder outputs clouds of $1024$ points, and the latent space dimension is $l=1024$. The regularization is done with balls of radius $0.1$ and the loss weighting $\lambda$ is set to $1e^{-2}$. Each gradient descent is performed until the loss does not improve for $100$ steps. The learning rate is $1e^{-2}$ at start, and is divided by 10 every time the loss doesn't improve for 10 consecutive steps.

\section{Experiments}
\label{sec:experiments}

\begin{figure}
    \centering
    \includegraphics[width=\linewidth]{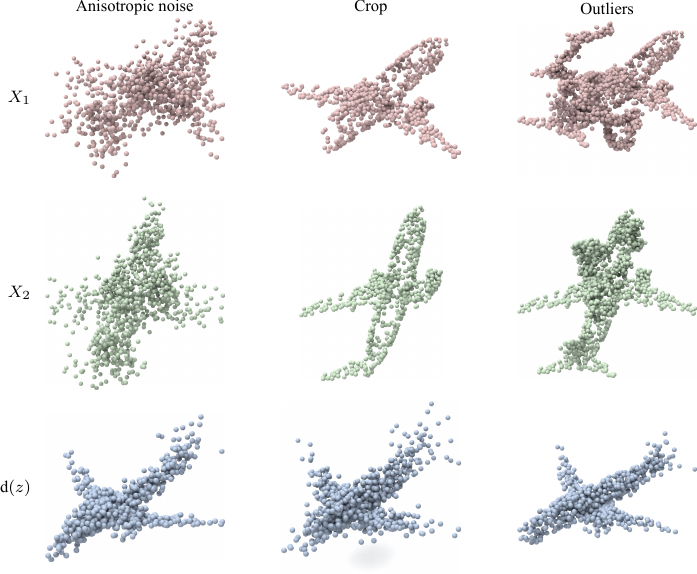}
    \caption{\textbf{Visual results}. First two rows: Degraded views to register. Last row: Template obtained after the optimization of the criterion $\Loss^f$ (\cref{eq:final_latent_optim}) with known visibility ratio $r=0.2$, outliers ratio $o=0.3$, and covariance matrix $\Sigma = \operatorname{diag}(0.03, 0.03, 0.15)$. Each optimization is done for $N = 100$ views (2 of which are shown). For each degradation, the estimated template $\D(z)$ has compensated for the degradation: it is complete, without outliers, and denoised.}
    \label{fig:viz}
\end{figure}
\subsection{Experimental protocol}
\paragraph{Baselines}
We compare POLAR with different classes of methods: (i) non-learning-based methods, which can be either pairwise, like FGR \cite{fgr_Zhou16} and more recently MAC \cite{mac_zhang23}, or based on a generative multiview paradigm, like JRMPC \cite{jrmpc_Evangelidis17} and EMPMR \cite{scam_Zhu20}; (ii) deep learning-based methods, which are pairwise, namely PointNetLK \cite{pnlk_Aoki19}, DCP \cite{dcp_Wang19}, RPM-Net \cite{rpmnet_Yew20}, DeepGMR \cite{deepgmr_Yan20}, or more recently GeoT \cite{Qin2022GeometricTF} and RoITr \cite{roitr_Yu22}, with the exception of SGHR \cite{robmulti_Wang23}, a recent deep learning based multiview method. For pairwise methods, we use a synchronization algorithm to retrieve multiview absolute poses. We selected the method described in \cite{mpls_Shi20}, which we empirically observed to provide the best results. It is based on an iterative reweighted least squares method coupled with a message passing algorithm that estimates corruption levels.

\paragraph{Data}
We use ModelNet40 \cite{shapenet_Song15}, a comprehensive CAD model repository of $4602$ objects spanning 40 categories as the synthetic dataset of reference. We follow the official training and validation split from \cite{shapenet_Song15}. Learning-based methods, including POLAR's autoencoder are trained on this training set. Following the usual processing \cite{rpmnet_Yew20, Qin2022GeometricTF}, symmetric classes are removed. All subsequent experiments are conducted on the obtained validation set, considering $100$ views for each registration task. Each view is obtained following the preprocessing of \cref{sec:ae_training} followed by the application of a degradation.
In all experiments involving degradations, we suppose that their parameters are known (namely the noise covariance matrix $\Sigma$, the visibility ratio $v$ and the outliers ratio $o$). The impact of fixing wrong values to these parameters is moderate and will be studied in \cref{sec:polar_study}.c. \amend{Since POLAR is tailored for object-level registration, motivated by the SMLM application (registration of many severely noised versions of an object), we do not consider scene-level datasets. POLAR is not optimally suited for scene registration for two reasons. First, in POLAR, the views are registered on an estimated template, hence requiring a global latent vector to encode the entire scene. Our autoencoder is able to accurately represent single objects, but its representational capacity is not sufficient to capture the details of large-scale scenes. Second, in POLAR, the template is initialized with one of the views. This can be problematic in large-scale scenes, where a single fragment represents only a small portion of the complete scene. As a result, the optimization is more likely to fall into a local minimum. In contrast, at the object level, each view is an occluded observation of the same complete object, leading to a higher overlap with the template. In \cref{sec:real_data}, we provide results on challenging object-level real data, namely the Faust partial dataset \cite{Bojanic24} and SMLM data \cite{dstorm08}.}
\paragraph{Metrics}
Regardless of the approach employed, a rigid transformation is obtained for each point cloud, which enables their alignment in a shared reference frame. Since the pose in this frame can be arbitrary, we independently examine the $N(N-1)/2$ relative poses between the pairs of views. The one related to $X_i$ and $X_j$ is computed from the estimated and ground truth rotations, respectively $\hat{\Rot}$ and $\Rot^\star$ as $\theta_{ij} = \angle \left(\hat{\Rot}_i \Rot_i^\star, \hat{\Rot}_j \Rot_j^\star\right)$. This is usually called Relative Rotation Error (RRE). In our experiments, we report the registration success rate, often called Registration Recall (RR), defined as the proportion of angles $\theta_{ij} $ that are below a threshold. We also show the cumulative distribution function (CDF) of $\theta_{ij}$, which provides a complete view of the registration performance, independent from the choice of a threshold (the registration recall for one threshold $t$ is a point of the CDF). In particular, the CDF allows us to distinguish the notions of {\it accuracy}, which corresponds to low values of $\theta_{ij}$ and reflects the quality of the alignment when the registration is successful, and {\it robustness}, which corresponds to higher values of $\theta_{ij}$ and denotes the ability to obtain coarse successful registration in challenging scenarios. In the context of global registration, we are more interested in {\it robustness}, since {\it accuracy} can be achieved by subsequent refinement. Translation parameters are not considered for evaluation because we observed that a proper estimation of the rotation consistently leads to a proper estimation of the translation.
\subsection{Comparison with other methods}
\paragraph{Ability to handle large transformations}
\begin{table}[t]
\caption{Registration Recall (RR, $t=\ang{10}$) evolution with initial angle. Global methods are shown in bold.}
\label{tab:initial_angle}
\vspace{-0.5cm}
\begin{center}
\begin{small}
\begin{sc}
\begin{tabular}{lcccc}
\hline
\abovespace\belowspace
Method & $\theta \leq \frac{\pi}{4}$ & $\theta \in \left[\frac{\pi}{4}, \frac{\pi}{2}\right]$ & $\theta \in \left[\frac{\pi}{2}, \frac{3\pi}{4}\right]$ & $\theta \in \left[\frac{3\pi}{4}, \pi\right]$\\
\hline
\abovespace
\textbf{FGR} & \textbf{99.46} & \textbf{97.9} & \textbf{96.67} & \textbf{95.9}\\
\textbf{MAC} & \textbf{99.46} & \textbf{97.29} & \textbf{93.19} & \textbf{92}\\
FMR & 96.05 & 80.18 & 51.88 & 32\\
JRMPC & 96.05 & 80.18 & 51.88 & 32\\
EMPMR & 69.21 & 10.14 & 0.28 & 0\\
PNLK & 76.57 & 31.56 & 3.68 & 0\\
DCP & 71.53 & 24.73 & 8.16 & 3\\
\textbf{RPM-Net} & \textbf{99.73} & \textbf{99.81} & \textbf{99.37} & \textbf{99}\\
\textbf{DeepGMR} & \textbf{99.59} & \textbf{98.18} & \textbf{97.29} & \textbf{98}\\
\textbf{GeoTr} & \textbf{99.80} & \textbf{99.42} & \textbf{99.00} & \textbf{98.66}\\
\textbf{RoITr} & \textbf{100} & \textbf{100} & \textbf{100} & \textbf{100}\\
\textbf{SGHR} & \textbf{100} & \textbf{100} & \textbf{100} & \textbf{100}\\
\belowspace
\textbf{POLAR} & \textbf{100} & \textbf{100} & \textbf{100} & \textbf{100}\\
\hline
\end{tabular}
\end{sc}
\end{small}
\end{center}
\vskip -0.1in
\end{table}

We first study the ability of selected methods to handle arbitrarily large transformations. We use the ModelNet40 validation set, preprocessed following the policy of \cref{sec:ae_training} (which include a uniformly random pose over the whole $\rotgroup$ group), without degradation. Once $N$ absolute poses have been estimated, the RRE and initial angle are computed for each pair. Finally, the RR is computed for four ranges of initial angles $\angle \left(\Rot_i^\star, \Rot_j^\star\right)$: $\leq \ang{45}, \in \left[\ang{45}, \ang{90}\right], \in \left[\ang{90}, \ang{135}\right], \in \left[\ang{135}, \ang{180}\right]$. The results are presented in \cref{tab:initial_angle}. Global methods are shown in bold. As advertised, methods based on the EM algorithm (JMPRC and EMPMR) can only converge locally. As for deep learning based methods, PointNetLK is also local due to the use of the Lucas Kanade algorithm. \amend{DCP relies on an SVD algorithm to retrieve a transformation from correspondences estimated through an attention map between pose variant features, hence it remains local. In subsequent experiments, these local methods are not considered. On the other hand, FGR and MAC are almost global, even if for large angles they may be fooled by near-symmetries, while RPM-Net, DeepGMR, GeoT, RoITr, SGHR, and POLAR maintain near perfect performances regardless of the initial angle.}
\paragraph{Isotropic Noise}
\begin{figure}
    \centering
        \includegraphics[width=\linewidth]{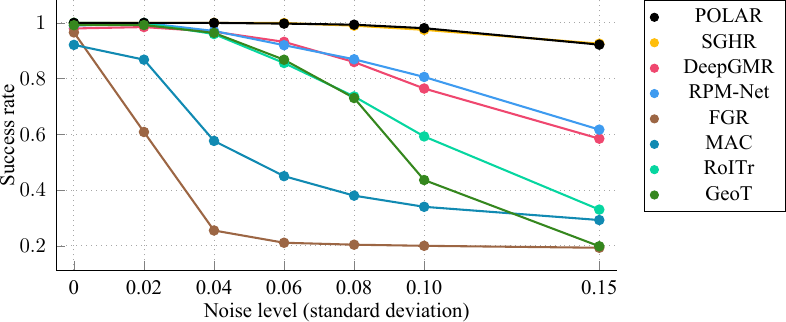}
    \caption{Registration Recall evolution with noise standard deviation.}
    \label{fig:experiment_isotropic_noise}
\end{figure}
We then study the robustness of these global methods to isotropic additive Gaussian noise. We consider increasing levels of noise following $\mathcal{N}(0, \sigma)$ with $\sigma$ ranging from $0$ to $0.15$. For one given noise level, each view is uniquely degraded by adding noise. The Registration recall is then computed for each level of noise (\cref{fig:experiment_isotropic_noise}). Methods purely based on local correspondences (FGR, MAC) struggle to handle large noise levels and as such, we do not consider them in further experiments. \amend{Even with their more robust correspondences estimation mechanisms, RoITr and GeoT fail due to the large local dissimilarity induced by strong noise.} It should be noted that POLAR and SGHR exhibit a strong robustness to isotropic noise.
\paragraph{Anisotropic Noise}   
\begin{figure*}
    \centering
    \includegraphics[width=\linewidth]{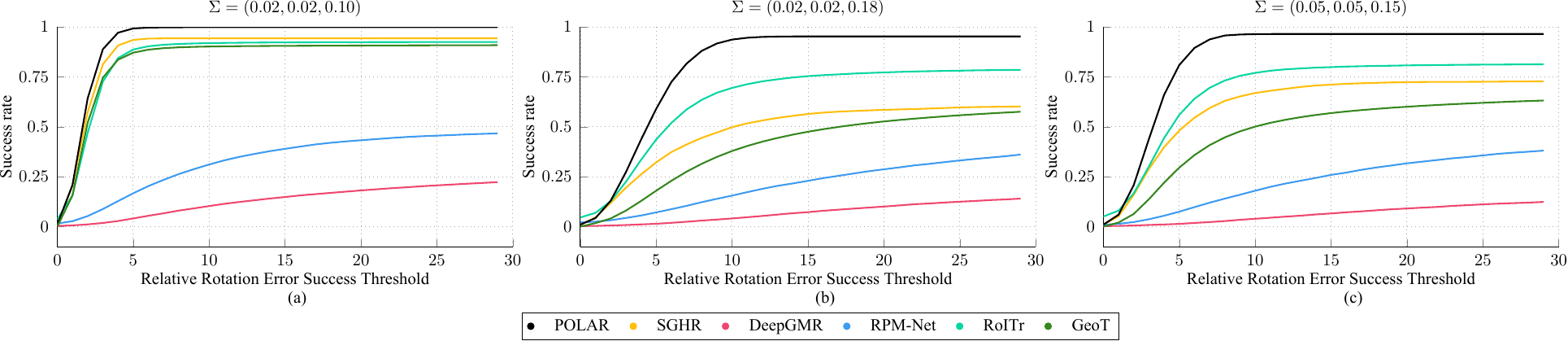}
    \caption{Cumulative distribution function of the registration recall for three anisotropic noises. (a) Low noise, low anisotropy. (b) Low noise, high anisotropy. (c) High noise, low anisotropy.}
    \label{fig:experiment_anisotropic_noise}
\end{figure*}%
A far harder degradation is the case of anisotropic noise, as it deforms each view in a unique way depending on its orientation. Two examples of such views are shown in \cref{fig:viz}. This often occurs in microscopic acquisition, in which the resolution in the microscope's axis is far lower than in its orthogonal plane. The greater the anisotropy of resolution, the greater the deformation undergone by objects. We study the impact of this anisotropy factor on   registration performances, measured by the full Registration Recall CDF, reported in \cref{fig:experiment_anisotropic_noise}. Under such a degradation, views are no longer locally similar. Therefore, methods purely based on local correspondences such as DeepGMR and RPM-Net cannot handle such cases. When the anisotropy factor and the noise level are relatively low, RoITr, GeoT, SGHR, and POLAR obtain good performances (\cref{fig:experiment_anisotropic_noise}.a). With the same noise level but a stronger anisotropy factor, performances decrease and only POLAR maintains a high success rate (\cref{fig:experiment_anisotropic_noise}.b). Similarly, with a relatively low anisotropy factor but stronger base noise level, POLAR maintains a high success rate when other methods performances drop (\cref{fig:experiment_anisotropic_noise}.c). An example of template obtained by POLAR is shown in \cref{fig:viz}.b. Compared to the input views, the reconstructed template is denoised and the anisotropy is corrected. To conclude, SGHR, RoITr and GeoT keep relatively good performances, as they partially leverage global descriptors: in SGHR, YOHO \cite{Wang_yoho22} features are processed by a NetVLAD \cite{Arandjelovic_netvlad18} module, while RoITr and GeoT gradually transforms local correspondences in more global ones using an attention mechanism. However, the POLAR approach obtains the best performances. 
\paragraph{Varying noise level}
\begin{figure}
    \centering
    \includegraphics[width=\linewidth]{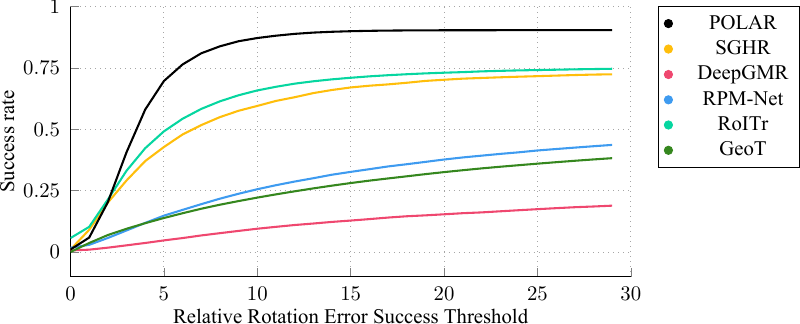}
    \caption{Cumulative distribution function of registration recall when registering views degraded by a varying noise level ($\sigma \sim \mathcal{U}\left(0.01, 0.2 \right)$).}
    \label{fig:experiment_varying_noise_level_per_object}
\end{figure}
To further study the ability of POLAR to be robust to local shape variations, we first consider the case of noisy data, with a different noise level for each view. Specifically, each view is noised with an isotropic Gaussian noise of variance $\sigma \sim \mathcal{U}\left(0.01, 0.2 \right)$. The results are shown in \cref{fig:experiment_varying_noise_level_per_object}. Methods based on local correspondences (DeepGMR, RPM-Net) are inherently limited, whereas POLAR maintains robust performance. SGHR and RoITr are partially based on global descriptors, hence they keep decent performances in this scenario. \amend{This experiment also highlights the benefit brought by the rotation-invariant cross-frame position awareness of RoITr over GeoT.}

\paragraph{Point density}
\begin{figure}
    \centering
    \includegraphics[width=\linewidth]{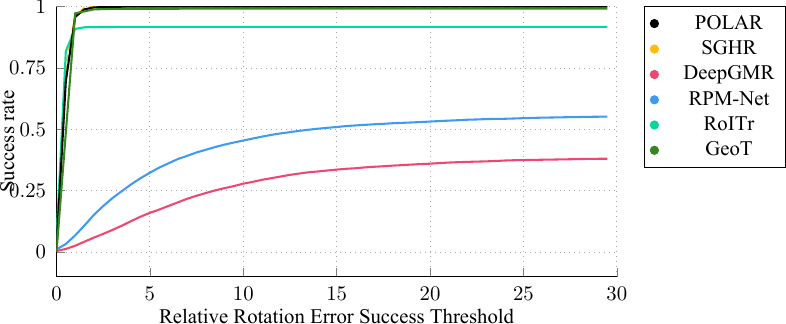}
    \caption{Cumulative distribution function of registration recall when registering views with varying point density ($n \sim \mathcal{U}\left(205, 1024 \right)$). The curve of SGHR is not visible because it is superimposed on the POLAR curve.}
    \label{fig:experiment_point_density}
\end{figure}
Similarly, we study the case where the views have different point densities. Specifically, the number of point of a view is randomly sampled in $\mathcal{U}\left(205, 1024 \right)$. The results are shown in \cref{fig:experiment_point_density}. POLAR is almost invariant to such degradation, and as for the varying noise level experiment, it obtains the best performances, on par with SGHR, closely followed by RoITr and GeoT, since they both partially leverage global descriptors.
\paragraph{Partial visibility}
\begin{figure}
    \centering
    \includegraphics[width=\linewidth]{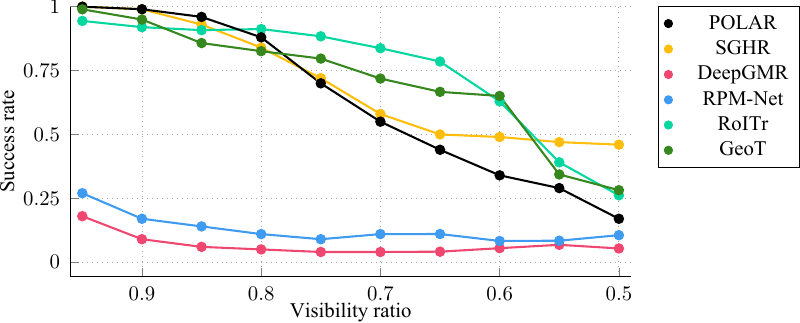}
    \caption{Evolution of the registration recall with the visibility ratio.}
    \label{fig:experiment_partial_visibility}
\end{figure}
Point cloud data often suffer from partial occlusion. We study the case of data where a fixed ratio of points has been cropped out for each view (two examples are shown in \cref{fig:viz}). The results are shown in \cref{fig:experiment_partial_visibility}. \amend{For high overlap, POLAR exhibits the best performance with SGHR, thanks to our occlusion handling masking in the loss (see \cref{eq:latent_optim_crop}). The coarse-to-fine learnable Sinkhorn from attention approach of RoITr and GeoT is the best algorithm for low overlaps up to a certain extent, at the price of a far greater computational complexity (studied in \cref{sec:experiments}.h). For even lower overlaps, SGHR is the best method, thanks to its NetVLAD module, specifically designed to handle such scenarios. These behaviors highlight specific design choices: the sparse pose graph initialized from local correspondences in SGHR is able to deal with very low overlaps as long as views remain locally similar, whereas the global descriptor of POLAR is suited to handle strong local dissimilarities, at the price of impaired performances for low overlaps. Indeed, the main limiting point of POLAR for low overlaps is the initialization of the template with the medoid of the encoded views (\cref{eq:medoid}), which gets increasingly incomplete with the occlusion.} An example of template reconstructed with POLAR is shown in \cref{fig:viz}. Despite the missing parts of the cropped view, the template is a complete plane.

\paragraph{Outliers}
\begin{figure}
    \centering
\includegraphics[width=\linewidth]{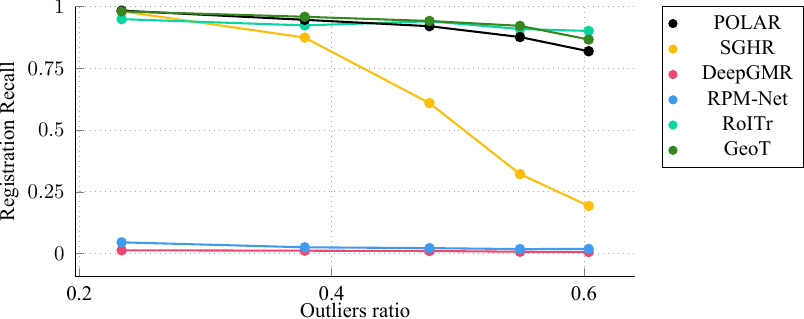}
    \caption{Evolution of the registration recall with the outliers ratio.}
    \label{fig:experiment_outliers}
\end{figure}
Finally, the last degradation considered is the presence of outliers. We simulate outliers by sampling points along a random curve starting from the object's surface, as can be seen on \cref{fig:viz}. This simulates the kind of outliers observed in SMLM data (\cref{sec:real_data}) and presents a more challenging case compared to usual simulations such as uniform sampling in the unit sphere, in which case outliers could be segmented out by a dedicated method easily. POLAR exhibits a strong robusteness to outliers, on par with that of RoITr and GeoT (\cref{fig:experiment_outliers}) which were specifically crafted to handle wrong correspondences. SGHR performances decrease gradually with the amount of outliers. DeepGMR and RPM-Net are not robust to outliers at all. An example of template obtained by POLAR is shown in \cref{fig:viz}. The template is free from outliers.
\paragraph{Time efficiency}
\begin{figure}
    \centering
    \includegraphics[width=\linewidth]{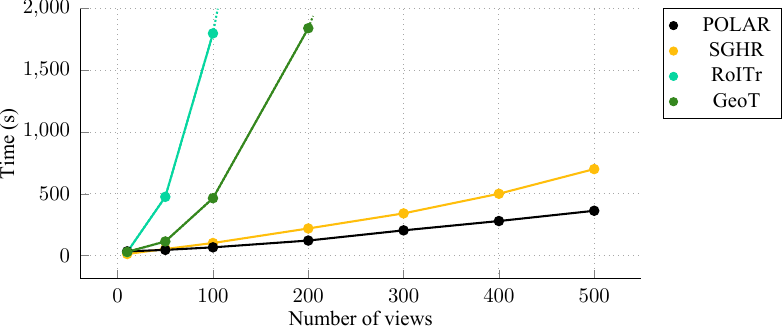}
    \caption{Evolution of the computation time with the number of views.}
    \label{fig:experiment_time_efficiency}
\end{figure}
We study the scalability of registration methods with the number of views for the three best performing methods. As pairwise methods leveraging big Transformer networks and approximated optimal transport, the computation cost of RoITr and GeoT increases very rapidly with the number of views. SGHR is also pairwise, but the NetVLAD module estimates the highest overlaps, and only the resulting subset of pairwise registrations is computed, thereby enhancing scalability. As a generative approach, POLAR scales linearly with the number of views, and yields the lowest computation cost.

\subsection{Study of POLAR} \label{sec:polar_study}

\paragraph{Number of views}
\begin{figure}
    \centering
    \includegraphics[width=\linewidth]{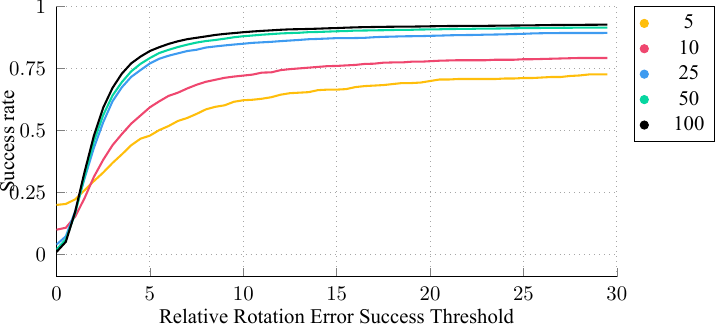}
    \caption{Evolution of the registration recall of POLAR with the number of views to register.}
    \label{fig:experiment_numviews}
\end{figure}
\begin{figure}
\centering
\subfloat[]{\includegraphics[width=0.2\textwidth]{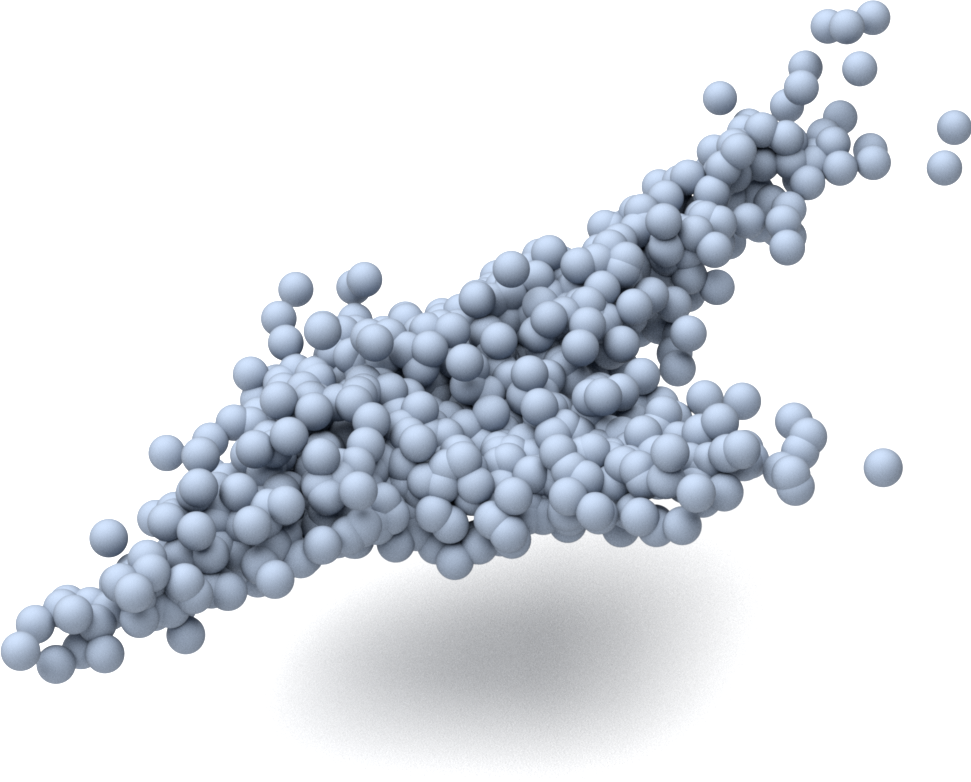}\label{fig:viz_numviews_5}}%
\subfloat[]{\includegraphics[width=0.2\textwidth]{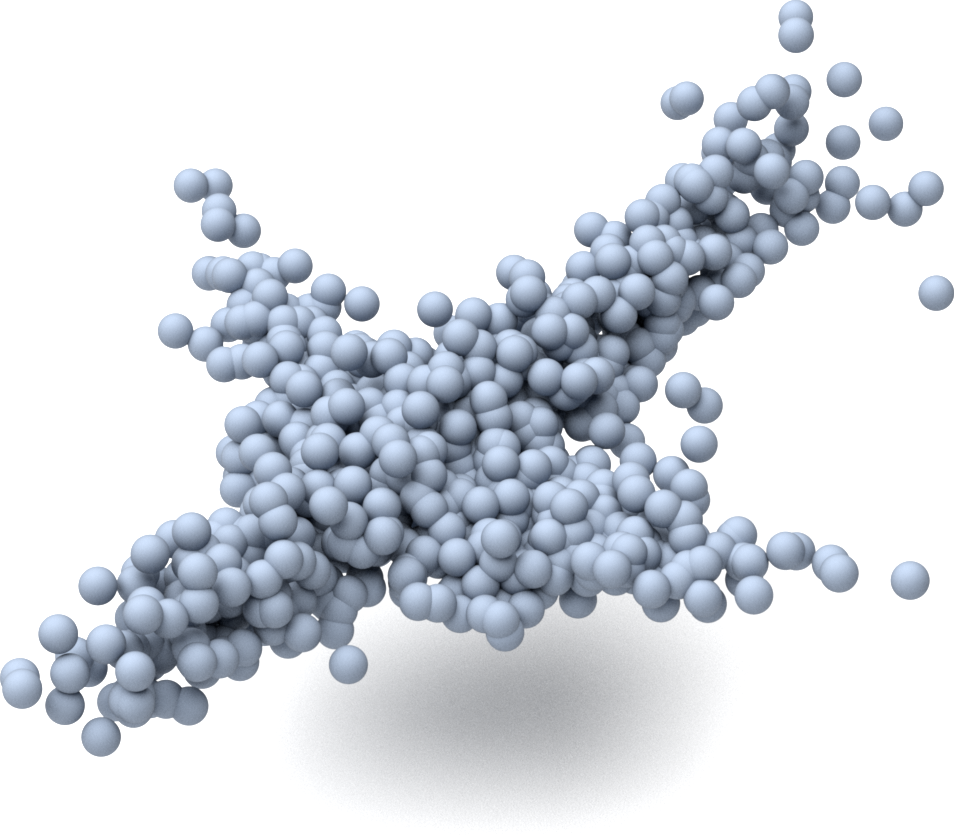} \label{fig:viz_numviews_100}}
\caption{Visual comparison of template reconstructed by POLAR from data combining the three degradations with (a) 5 and (b) 100 views.}
\label{fig:viz_numviews}
\end{figure}
We study the impact of the number of views on the registration performances. We generate 100 views combining three degradations (20\% of outliers, 20\% of occlusion, isotropic noise of standard deviation $0.02$). POLAR is optimized on the first $n$ views, with $n$ in $\{5, 10, 25, 50, 50\}$. The results are shown on \cref{fig:experiment_numviews}. Performances gradually increase with the number of views. POLAR benefits from an increasing number of views since it allows the template to be more accurately estimated. This is illustraded in \cref{fig:viz_numviews} where the estimated template for $n=5$ (\cref{fig:viz_numviews_5}) and $n=100$ (\cref{fig:viz_numviews_100}) are displayed. In the latter case, the reconstruction is complete and more precise.  

\paragraph{Ablation studies}\label{sec:ablation_studies}
\begin{figure*}
    \centering
    \includegraphics[width=\linewidth]{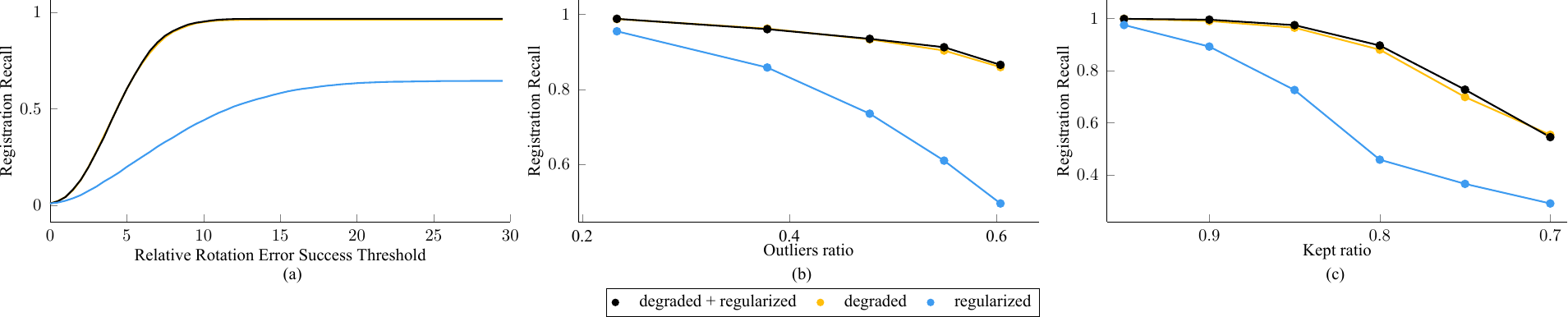}
    \caption{\textbf{Ablation studies on the three degradations.} \textbf{(a)} Cumulative distribution function of registration recall on data corrupted by an anisotropic noise of covariance $\Sigma = \operatorname{diag}\left(0.02, 0.02, 0.18\right)$. \textbf{(b)} Registration recall evolution on data gradually corrupted by outliers. \textbf{(c)} Registration recall evolution on data gradually cropped out. \textit{regularized} is the criterion in \cref{eq:latent_optim}, without the degradation model, but with the regularization of \cref{sec:latent_loss}.f. \textit{degraded} is the degraded criterion of \cref{eq:degraded_latent_optim}, but without the regularization. \textit{degraded $+$ regularized} is the criterion of \cref{eq:final_latent_optim} that combines both the degradation model and the regularization.}
    \label{fig:ablation_study}
\end{figure*}%
We now study the impact of two components of the criterion \eqref{eq:final_latent_optim}: the regularization, and the degradation model. The full version of the criterion, \ie the standard POLAR algorithm using the loss \eqref{eq:final_latent_optim}, is named \textit{degraded + regularized}. In order to examine the influence of the regularization term, we set $\lambda = 0$ in \eqref{eq:final_latent_optim}, obtaining the \textit{degraded} version. We study a third version of POLAR that we name \textit{regularized}, by keeping the regularization term but replacing in \eqref{eq:final_latent_optim} the degraded
criterion $\Loss(\bz, \Rho)$ by its basic version \eqref{eq:latent_optim}. These three variations of the optimization criterion are compared on data corrupted by the three types of degradation considered: anisotropic noise (\cref{fig:ablation_study}.a), occlusion (\cref{fig:ablation_study}.b), and outliers (\cref{fig:ablation_study}.c). We observe that the regularization has no impact from a pure registration performance perspective. It should be noted though, that obtained templates with regularization are visually more appealing to a human eye. For each of the addressed degradation, while the basic latent approach already brings decent performance, the degraded version of the criterion greatly enhances the algorithm's robustness.

\paragraph{Robustness to visibility and outlier ratios} \label{sec:robustness}
For the anisotropic case (\cref{fig:ablation_study}.a), the noise properties must be known. These properties can often be estimated separately, which is notably the case on SMLM data studied in \cref{sec:real_data}. However, estimating the ratio of outliers and occlusion is more difficult and limiting in practice. To evaluate the sensitivity of POLAR to these ratios, we study the impact of wrong ratio values on the registration performances. 

\begin{figure}
    \centering
\includegraphics[width=\linewidth]{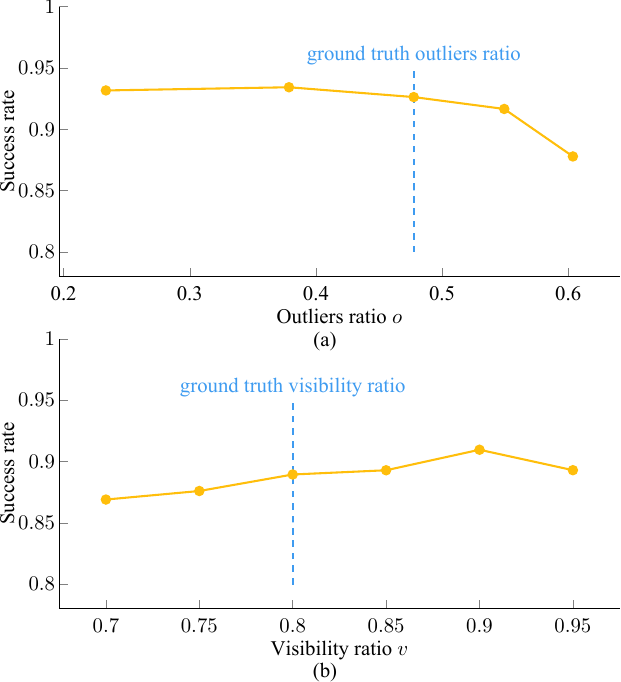}
    \caption{Registration recall evolution when using varying masking ratios to register the same data.}
    \label{fig:unknown_degradations}
\end{figure}

Specifically, we consider two cases: data with $20\%$ cropped out (\cref{fig:unknown_degradations}.a), and data with $\sim 50\%$ of outliers (\cref{fig:unknown_degradations}.b). We then run several registrations using varying values for the visibility ratio $v$ and the outliers ratio $o$. In both cases, a rough estimate of the ground truth ratio does not severely impact performances, and slightly underestimating the level of degradation improves results. This proves that POLAR is usable even when the visibility and outliers ratios are unknown, making it suitable for application where the degradation model is unknown. Moreover, an extension of the present method could consist in estimating this degradation model alongside the registration task. 

\subsection{Real Data} \label{sec:real_data}

\begin{figure}
    \centering
    \includegraphics[width=0.9\linewidth]{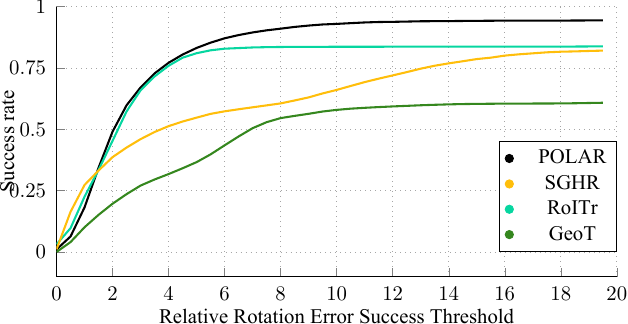}
    \caption{Cumulative distribution function of the registration recall on realistically occluded data from the Faust partial dataset \cite{Bojanic24}.}
    \label{fig:experiment_faust_partial}
\end{figure}

\paragraph{Realistic occlusion with FAUST-partial}
We now study the registration performances on data from the FAUST-partial dataset \cite{Bojanic24}. FAUST-partial is comprised of 100 human body scans. Realistic occlusions are generated by applying the hidden point removal algorithm on icosahedron points around each scan. On these data, each view has a specific visibility ratio. Thus, we select a subset of views from which this ratio is in $\left[0.7, 1\right]$ and run POLAR with a fixed ratio of $0.9$, which in most cases slightly underestimates the degradation, following the results from \cref{fig:unknown_degradations}.b. POLAR is the best method in that case (\cref{fig:experiment_faust_partial}). Note that this also highlights the generalization capability of POLAR. Indeed, the shapes here are unknown to the autoencoder. Hence, reconstructed templates are less precise. Nonetheless, the estimated poses are still correct.
\paragraph{Highly degraded SMLM data}
\begin{figure*}
\centering
\subfloat[]{\includegraphics[width=0.27\linewidth]{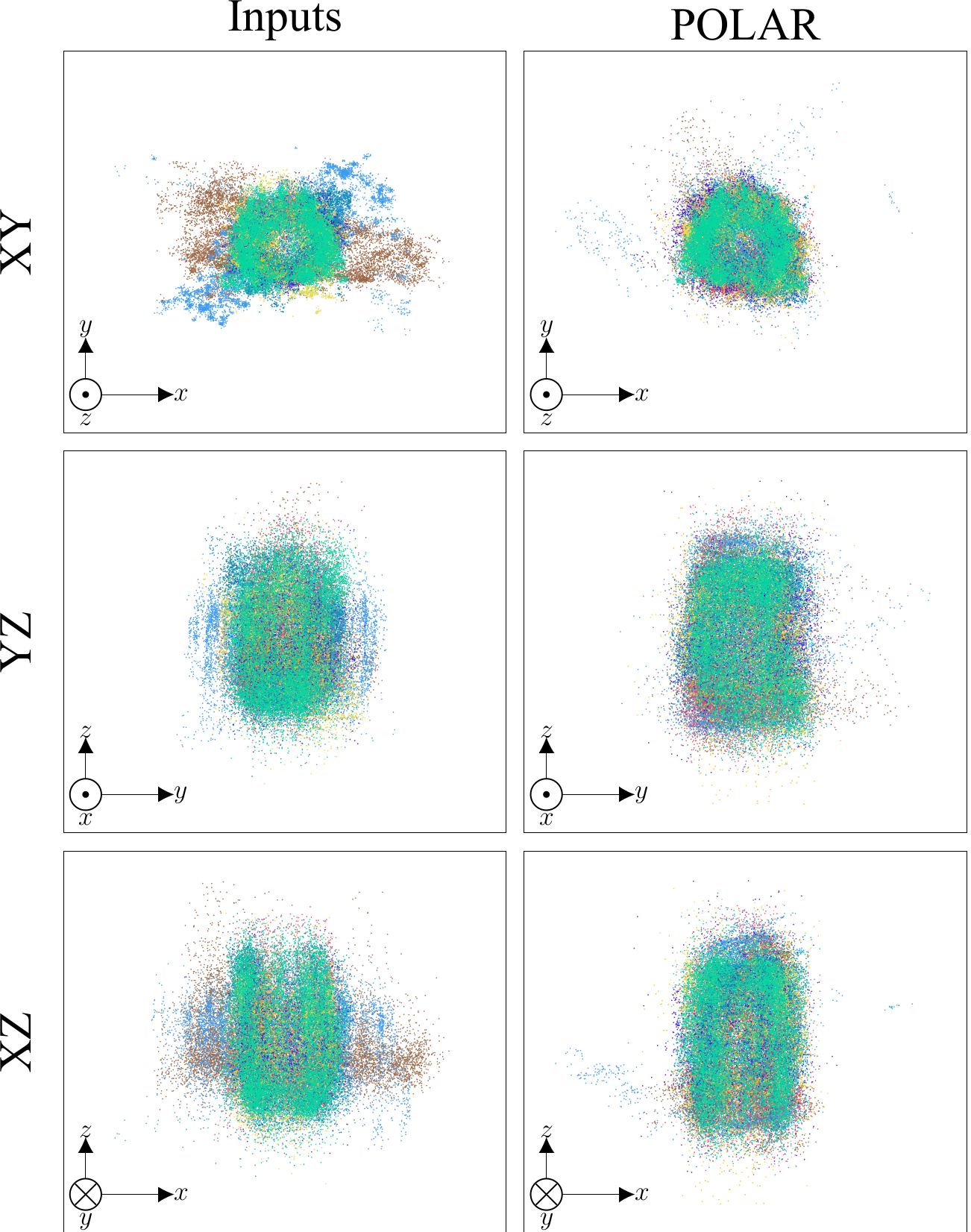}\label{fig:viz_dstorm_dataset}}%
\hfil
\subfloat[]{\includegraphics[width=0.635\linewidth]{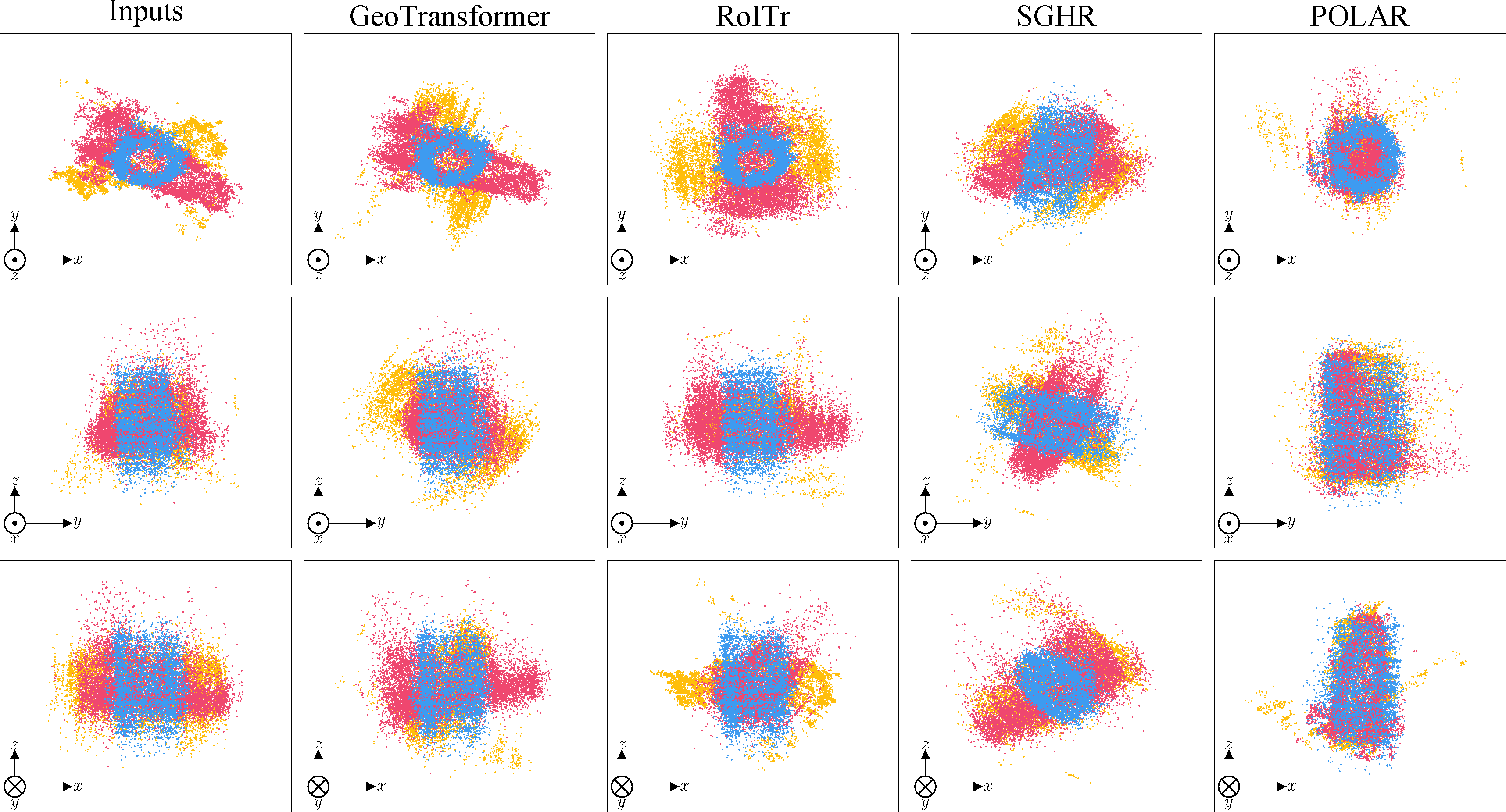} \label{fig:viz_dstorm_registration}}
\caption{\textbf{Visual comparison of registration results on SMLM data}. \textbf{(a)} Whole dataset of 9 centrioles, in initial poses (first column) and after registration with POLAR (second column). \textbf{(b)} Selected subset of three centrioles. In each line, the point clouds are observed in a different plane ($\mathrm{XY}$, $\mathrm{YZ}$ and $\mathrm{XZ}$). The resolution anisotropy can be seen in the inputs point clouds (columns $1$ and $3$): the resolution in the plane $\mathrm{XY}$ is much higher than in the planes $\mathrm{YZ}$ and $\mathrm{XZ}$.}
\end{figure*}
We use data obtained with direct optical reconstruction microscopy (dStorm) \cite{dstorm08}, combined with expansion microscopy. The data has been acquired in the group of Markus Sauer in University of Würzburg. It is composed of nine identical particles called centrioles, observed from different views. This is the most challenging dataset, as each view is heavily corrupted with strong anisotropic noise, outliers, missing parts. The SMLM acquisition process provides point clouds with a known 3D uncertainty $\Sigma$. The resolution along the microscope axis ($z$ in \cref{fig:viz_dstorm_dataset}) is much worse than in the orthogonal plane, which induces a large deformation of the object in the XZ and YZ planes. The shape of a centriole can be coarsely approximated by a cylinder. Hence, when it is aligned with the microscope axis, the cylinder shape is clearly visible (blue particle in \cref{fig:viz_dstorm_registration}) whereas other orientations result in a loss of this characteristic shape  due to the anisotropy of resolution. The centriole is often attached to tubular structure called microtubules, which can be considered as outliers (see the yellow particle in \cref{fig:viz_dstorm_registration}). Finally, SMLM data are usually corrupted by a high level of partial visibility, since each point corresponds to the fluorescent emission of a fluorophore, and the fluorophores do not cover uniformly the surface of the particle. Hence, this modality combines a high level of each of the three degradations addressed by POLAR. In \cref{fig:viz_dstorm_dataset}, we show the whole set of nine centrioles, alongside POLAR results. Then in \cref{fig:viz_dstorm_registration}, for the sake of clarity, we select a subset of three particles to visually compare all methods. Among all the tested methods, POLAR is the only one able to correctly register centrioles.

\section{Conclusion}\label{sec:conclusion}
We introduced Point Cloud Latent Registration (POLAR), an algorithm designed to simultaneously estimate a set of rigid motions that align numerous severely degraded views. POLAR integrates a global descriptor derived from a pretrained autoencoder, a global optimization framework, and an informed criterion taking degradations into account. By combining these elements, POLAR demonstrates state-of-the-art performance on both synthetic and real-world datasets that are significantly affected by anisotropic noise, partial visibility, and outliers.

\bibliographystyle{IEEEtran}
\bibliography{IEEEabrv,main}

\begin{thebibliography}{10}
\providecommand{\url}[1]{#1}
\csname url@samestyle\endcsname
\providecommand{\newblock}{\relax}
\providecommand{\bibinfo}[2]{#2}
\providecommand{\BIBentrySTDinterwordspacing}{\spaceskip=0pt\relax}
\providecommand{\BIBentryALTinterwordstretchfactor}{4}
\providecommand{\BIBentryALTinterwordspacing}{\spaceskip=\fontdimen2\font plus
\BIBentryALTinterwordstretchfactor\fontdimen3\font minus \fontdimen4\font\relax}
\providecommand{\BIBforeignlanguage}[2]{{%
\expandafter\ifx\csname l@#1\endcsname\relax
\typeout{** WARNING: IEEEtran.bst: No hyphenation pattern has been}%
\typeout{** loaded for the language `#1'. Using the pattern for}%
\typeout{** the default language instead.}%
\else
\language=\csname l@#1\endcsname
\fi
#2}}
\providecommand{\BIBdecl}{\relax}
\BIBdecl

\bibitem{Heydarian21}
H.~Heydarian, M.~Joosten, A.~Przybylski, F.~Schueder, R.~Jungmann, B.~v. Werkhoven, J.~Keller-Findeisen, J.~Ries, S.~Stallinga, M.~Bates \emph{et~al.}, ``3d particle averaging and detection of macromolecular symmetry in localization microscopy,'' \emph{Nature communications}, vol.~12, no.~1, p. 2847, 2021.

\bibitem{pcr_Blais95}
G.~Blais and M.~D. Levine, ``Registering multiview range data to create 3d computer objects,'' \emph{IEEE TPAMI}, 1995.

\bibitem{Nuchter07}
A.~N. \"uchter, K.~Lingemann, J.~Hertzberg, and H.~Surmann, ``6d slam–3d mapping outdoor environments,'' \emph{J. Field Robotics}, 2007.

\bibitem{fusion_Newcombe15}
R.~Newcombe, D.~Fox, and S.~Seitz, ``Dynamicfusion: Reconstruction and tracking of non-rigid scenes in real-time,'' \emph{CVPR}, 2015.

\bibitem{kinectfusion_Newcombe15}
R.~A. Newcombe, S.~Izadi, O.~Hilliges, D.~Molyneaux, D.~Kim, A.~J. Davison, P.~Kohi, J.~Shotton, S.~Hodges, and A.~Fitzgibbon, ``Kinectfusion: Real-time dense surface mapping and tracking,'' \emph{IEEE International Symposium on Mixed and Augmented Reality}, 2011.

\bibitem{4dreconstruct_Tang21}
J.~Tang, D.~Xu, K.~Jia, and L.~Zhang, ``Learning parallel dense correspondence from spatio-temporal descriptors for efficient and robust 4d reconstruction,'' \emph{CVPR}, 2021.

\bibitem{hybridmixture_Min20}
Z.~Min, J.~Wang, and M.~Q.~H. Meng, ``Robust generalized point cloud registration using hybrid mixture model,'' \emph{International Conference on Robotics and Automation (ICRA)}, 2018.

\bibitem{globmo_Govindu18}
A.~Chatterjee and V.~Govindu, ``Robust relative rotation averagings,'' \emph{IEEE TPAMI}, 2018.

\bibitem{globmo_Torsello11}
A.~Torsello, E.~Rodola, and A.~Albarelli, ``Multiview registration via graph diffusion of dual quaternions,'' \emph{CVPR}, 2011.

\bibitem{globmo_Maset17}
E.~Maset, F.~Arrigoni, and A.~Fusiello, ``Practical and efficient multi-view matching,'' \emph{ICCV}, 2017.

\bibitem{globmo_Birdal18}
T.~Birdal, U.~Simsekli, M.~O. Eken, and S.~Ilic, ``Bayesian pose graph optimization via bingham distributions and tempered geodesic mcmc,'' \emph{NeurIPS}, 2018.

\bibitem{globmo_Bernard15}
F.~Bernard, J.~Thunberg, P.~Gemmar, F.~Hertel, A.~Husch, and J.~Goncalves, ``A solution for multi-alignment by transformation synchronisation,'' \emph{CVPR}, 2015.

\bibitem{globmo_Arrigoni16}
F.~Arrigoni, B.~Rossi, and A.~Fusiello, ``Spectral synchronization of multiple views in se(3),'' \emph{Journal on Imaging Sciences}, 2016.

\bibitem{globmo_Arrigoni14}
F.~Arrigoni, L.~Magri, B.~Rossi, P.~Fragneto, and A.~Fusiello, ``Robust absolute rotation estimation via low-rank and sparse matrix decomposition,'' \emph{IEEE International Conference on 3D Vision (3DV)}, 2014.

\bibitem{globmo_ArieNachimson212}
M.~Arie-Nachimson, S.~Z. Kovalsky, I.~Kemelmacher-Shlizerman, A.~Singer, and R.~Basri, ``Global motion estimation from point matches,'' \emph{International Conference on 3D Imaging, Modeling, Processing, Visualization and Transmission}, 2012.

\bibitem{robmulti_Wang23}
H.~Wang, Y.~Liu, Z.~Dong, Y.~Guo, Y.-S. Liu, W.~Wang, and B.~Yang, ``Robust multiview point cloud registration with reliable pose graph initialization and history reweighting,'' \emph{CVPR}, 2023.

\bibitem{Jian11}
B.~Jian and B.~C. Vemuri, ``Robust point set registration using {Gaussian} mixture models,'' \emph{IEEE TPAMI}, 2011.

\bibitem{mlmd_Eckart15}
B.~Eckart, K.~Kim, A.~Troccoli, A.~Kelly, and J.~Kautz, ``Mlmd: Maximum likelihood mixture decoupling for fast and accurate point cloud registration,'' \emph{International Conference on 3D Vision}, 2015.

\bibitem{em_Chui20}
H.~Chui and A.~Rangarajan, ``A feature registration framework using mixture models,'' \emph{IEEE Workshop on Mathematical Methods in Biomedical Image Analysis}, 2000.

\bibitem{jrmpc_Evangelidis17}
G.~D. Evangelidis and R.~Horaud, ``Joint alignment of multiple point sets with batch and incremental expectation-maximization,'' \emph{IEEE TPAMI}, 2017.

\bibitem{cpd_Myronenko10}
A.~Myronenko and X.~Song, ``Point set registration : Coherent point drift,'' \emph{IEEE TPAMI}, 2010.

\bibitem{filterreg_Gao19}
W.~Gao and R.~Tedrake, ``Filterreg : Robust and efficient probabilistic point-set registration using gaussian filter and twist parameterization,'' \emph{CVPR}, 2019.

\bibitem{scam_Zhu20}
J.~Zhu, R.~Guo, Z.~Li, J.~Zhang, and S.~Pang, ``Registration of multi-view point sets under the perspective of expectation-maximization,'' \emph{IEEE TIP}, 2020.

\bibitem{tstudent_Ma22}
Y.~Ma, J.~Zhu, Z.~Tian, and Z.~Li, ``Effective multiview registration of point clouds based on student’s-t mixture model,'' \emph{Information Sciences}, 2022.

\bibitem{lieavg_Govindu2004}
V.~Govindu, ``Lie-algebraic averaging for globally consistent motion estimation,'' \emph{CVPR}, 2004.

\bibitem{synchrolearn_Huang19}
X.~Huang, Z.~Liang, X.~Zhou, Y.~Xie, L.~Guibas, and Q.~Huang, ``Learning transformation synchronization,'' \emph{CVPR}, 2019.

\bibitem{deepmapping_Ding19}
L.~Ding and C.~Feng, ``Deepmapping: Unsupervised map estimation from multiple point clouds,'' \emph{CVPR}, 2019.

\bibitem{learn_irts_Yew21}
Z.~J. Yew and G.~H. Lee, ``Learning iterative robust transformation synchronization,'' \emph{2021 International Conference on 3D Vision (3DV)}, 2021.

\bibitem{learnmulti_Gojcic20}
Z.~Gojcic, C.~Zhou, J.~D. Wegner, L.~J. Guibas, and T.~Birdal, ``Learning multiview 3d point cloud registration,'' \emph{CVPR}, 2020.

\bibitem{robustMR_Zhang21}
J.~Zhang, M.~Zhao, X.~Jiang, and D.~Yan, ``Robust multi-view registration of point sets with laplacian mixture model,'' \emph{ACCV}, 2021.

\bibitem{fuzzyCG_Liao22}
Q.~Liao, D.~wei Sun, S.~Zhang, A.~Loutfi, and H.~Andreasson, ``Fuzzy cluster-based group-wise point set registration with quality assessment,'' \emph{IEEE TIP}, 2022.

\bibitem{infoGM_Wang22}
W.~Wang and K.~Lin, ``Information granule-based multi-view point sets registration using fuzzy c-means clustering,'' \emph{Multimedia Tools and Applications}, 2022.

\bibitem{icp_Besl92}
P.~J. Besl and N.~McKay, ``A method for registration of 3-d shapes,'' \emph{IEEE Trans. on Pattern Analysis and Machine Intelligence}, 1992.

\bibitem{icpvariant_Rusink01}
S.~Rusinkiewicz and M.~Levoy, ``Efficient variants of the icp algorithm,'' \emph{Third International Conference on 3-D Digital Imaging and Modeling}, 2001.

\bibitem{Ao2020SpinNetLA}
S.~Ao, Q.~Hu, B.~Yang, A.~Markham, and Y.~Guo, ``Spinnet: Learning a general surface descriptor for 3d point cloud registration,'' \emph{CVPR}, 2021.

\bibitem{Deng18}
H.~Deng, T.~Birdal, , and S.~Ilic, ``Ppf-foldnet: Unsupervised learning of rotation invariant 3d local descriptors,'' \emph{ECCV}, 2018.

\bibitem{Yu2022RIGARA}
H.~Yu, J.~Hou, Z.~Qin, M.~Saleh, I.~S. Shugurov, K.~Wang, B.~Busam, and S.~Ilic, ``Riga: Rotation-invariant and globally-aware descriptors for point cloud registration,'' \emph{IEEE TPAMI}, 2022.

\bibitem{roitr_Yu22}
H.~Yu, Z.~Qin, J.~Hou, M.~Saleh, D.~Li, B.~Busam, and S.~Ilic, ``Rotation-invariant transformer for point cloud matching,'' \emph{CVPR}, 2023.

\bibitem{Qin2022GeometricTF}
Z.~Qin, H.~Yu, C.~Wang, Y.~Guo, Y.~Peng, and K.~Xu, ``Geometric transformer for fast and robust point cloud registration,'' \emph{CVPR}, 2022.

\bibitem{Bai2020D3FeatJL}
X.~Bai, Z.~Luo, L.~Zhou, H.~Fu, L.~Quan, and C.-L. Tai, ``D3feat: Joint learning of dense detection and description of 3d local features,'' \emph{CVPR}, 2020.

\bibitem{Saleh2020GraphiteGF}
M.~Saleh, S.~Dehghani, B.~Busam, N.~Navab, and F.~Tombari, ``Graphite: Graph-induced feature extraction for point cloud registration,'' \emph{International Conference on 3D Vision (3DV)}, 2020.

\bibitem{perfectmatch_Gojcic19}
Z.~Gojcic, C.~Zhou, J.~D. Wegner, and A.~Wieser, ``The perfect match: 3d point cloud matching with smoothed densities,'' \emph{CVPR}, 2019.

\bibitem{evomultidesc_Wu24}
Y.~Wu, J.~Sheng, H.~Ding, P.~Gong, H.~Li, M.~Gong, W.~Ma, and Q.~Miao, ``Evolutionary multitasking descriptor optimization for point cloud registration,'' \emph{IEEE Transactions on Evolutionary Computation}, 2024.

\bibitem{egst_Yuan24}
Y.~Yuan, Y.~Wu, X.~Fan, M.~Gong, W.~Ma, and Q.~Miao, ``Egst: Enhanced geometric structure transformer for point cloud registration,'' \emph{IEEE Transactions on Visualization and Computer Graphics}, 2024.

\bibitem{Yew2022REGTREP}
Z.~J. Yew and G.~H. Lee, ``Regtr: End-to-end point cloud correspondences with transformers,'' \emph{CVPR}, 2022.

\bibitem{mpct_Wu24}
Y.~Wu, J.~Liu, M.~Gong, Z.~Liu, Q.~Miao, and W.~Ma, ``Mpct: Multiscale point cloud transformer with a residual network,'' \emph{IEEE Transactions on Multimedia}, 2024.

\bibitem{fcgf_Choy19}
C.~Choy, J.~Park, and V.~Koltun, ``Fully convolutional geometric features,'' \emph{ICCV}, 2019.

\bibitem{Deng2018PPFNetGC}
H.~Deng, T.~Birdal, and S.~Ilic, ``Ppfnet: Global context aware local features for robust 3d point matching,'' \emph{CVPR}, 2018.

\bibitem{Huang2020PREDATORRO}
S.~Huang, Z.~Gojcic, M.~M. Usvyatsov, A.~Wieser, and K.~Schindler, ``Predator: Registration of 3d point clouds with low overlap,'' \emph{CVPR}, 2021.

\bibitem{Li2021LepardLP}
Y.~Li and T.~Harada, ``Lepard: Learning partial point cloud matching in rigid and deformable scenes,'' \emph{CVPR}, 2021.

\bibitem{Saleh2022BendingGH}
M.~Saleh, S.~cheng Wu, L.~D. Cosmo, N.~Navab, B.~Busam, and F.~Tombari, ``Bending graphs: Hierarchical shape matching using gated optimal transport,'' \emph{CVPR}, 2022.

\bibitem{Yu2021CoFiNetRC}
H.~Yu, F.~Li, M.~Saleh, B.~Busam, and S.~Ilic, ``Cofinet: Reliable coarse-to-fine correspondences for robust point cloud registration,'' \emph{NeurIPS}, 2021.

\bibitem{3dmatch_Zeng17}
A.~Zeng, S.~Song, M.~Nießner, M.~Fisher, J.~Xiao, and T.~Funkhouser, ``3dmatch : Learning local geometric descriptors from rgb-d reconstructions,'' \emph{CVPR}, 2017.

\bibitem{Zhang2023PCRCGPC}
Y.~Zhang, J.~Yu, X.~Huang, W.~Zhou, and J.~Hou, ``Pcr-cg: Point cloud registration via deep explicit color and geometry,'' in \emph{ECCV}, 2023.

\bibitem{ransac_Fischler81}
M.~A. Fischler and R.~C. Bolles, ``Random sample consensus: a paradigm for model fitting with applications to image analysis and automated cartography,'' \emph{COMMUN ACM}, 1981.

\bibitem{Barth2017GraphCutR}
D.~Bar{\'a}th and J.~Matas, ``Graph-cut ransac,'' \emph{CVPR}, 2017.

\bibitem{Quan2020CompatibilityGuidedSC}
S.~Quan and J.~Yang, ``Compatibility-guided sampling consensus for 3-d point cloud registration,'' \emph{IEEE Transactions on Geoscience and Remote Sensing}, 2020.

\bibitem{Rusu2009FastPF}
R.~B. Rusu, N.~Blodow, and M.~Beetz, ``Fast point feature histograms (fpfh) for 3d registration,'' \emph{IEEE International Conference on Robotics and Automation (ICRA)}, 2009.

\bibitem{Yang2022CorrespondenceSW}
J.~Yang, J.~Chen, S.~Quan, W.~Wang, and Y.~Zhang, ``Correspondence selection with loose–tight geometric voting for 3-d point cloud registration,'' \emph{IEEE Transactions on Geoscience and Remote Sensing}, 2022.

\bibitem{Yang2021SACCOTSC}
J.~Yang, Z.~Huang, S.~Quan, Z.~Qi, and Y.~Zhang, ``Sac-cot: Sample consensus by sampling compatibility triangles in graphs for 3-d point cloud registration,'' \emph{IEEE Transactions on Geoscience and Remote Sensing}, 2021.

\bibitem{Yang2021TowardEA}
J.~Yang, Z.~Huang, S.~Quan, Q.~Zhang, Y.~Zhang, and Z.~Cao, ``Toward efficient and robust metrics for ransac hypotheses and 3d rigid registration,'' \emph{IEEE Transactions on Circuits and Systems for Video Technology}, 2021.

\bibitem{pnlk_Aoki19}
Y.~Aoki, H.~Goforth, R.~A. Srivatsan, and S.~Lucey, ``Pointnetlk: Robust and efficient point cloud registration using pointnet,'' \emph{CVPR}, 2019.

\bibitem{pcrnet_Sarode19}
V.~Sarode, X.~Li, H.~Goforth, Y.~Aoki, R.~A. Srivatsan, S.~Lucey, and H.~Choset, ``Pcrnet: Point cloud registration network using pointnet encoding,'' in \emph{ArXiv e-prints}, 2019.

\bibitem{fmr_Huang2020}
X.~Huang, G.~Mei, and J.~Zhang, ``Feature-metric registration: A fast semi-supervised approach for robust point cloud registration without correspondences,'' \emph{CVPR}, 2020.

\bibitem{omnet_Xu21}
H.~Xu, S.~Liu, G.~Wang, G.~Liu, and B.~Zeng, ``Omnet: Learning overlapping mask for partial-to-partial point cloud registration,'' \emph{ICCV}, 2021.

\bibitem{Zhu2022CorrespondenceFreePC}
M.~Zhu, M.~Ghaffari, and H.~Peng, ``Correspondence-free point cloud registration with so(3)-equivariant implicit shape representations,'' \emph{Conference on Robot Learning}, 2022.

\bibitem{goicp_Yang16}
J.~Yang, H.~Li, D.~Campbell, and Y.~Jia, ``Go-icp: A globally optimal solution to 3d icp point-set registration,'' \emph{IEEE TPAMI}, 2016.

\bibitem{gradutednonconvexity_Yang16}
H.~Yang, P.~Antonante, V.~Tzoumas, and L.~Carlone, ``Graduated non-convexity for robust spatial perception: From non-minimal solvers to global outlier rejection,'' \emph{International Conference on Robotics and Automation (ICRA)}, 2019.

\bibitem{fgr_Zhou16}
Q.-Y. Zhou, J.~Park, and V.~Koltun, ``Fast global registration,'' \emph{ECCV}, 2016.

\bibitem{Hartley07}
R.~I. Hartley and F.~Kahl, ``Global optimization through searching rotation space and optimal estimation of the essential matrix,'' \emph{ICCV}, 2007.

\bibitem{Enqvist08}
O.~Enqvist and F.~Kahl, ``Robust optimal pose estimation,'' \emph{ECCV}, 2008.

\bibitem{Olsson09}
C.~Olsson, F.~Kahl, and M.~Oskarsson, ``Branch-and-bound methods for euclidean registration problems,'' \emph{IEEE TPAMI}, 2009.

\bibitem{deepgmr_Yan20}
W.~Yuan, B.~Eckart, K.~Kim, V.~Jampani, D.~Fox, and J.~Kautz, ``Deepgmr : Learning latent gaussian mixture models for registration,'' \emph{ECCV}, 2020.

\bibitem{Xie20}
S.~Xie, J.~Gu, D.~Guo, C.~R. Qi, L.~Guibas, and O.~Litany, ``Pointcontrast: Unsupervised pretraining for 3d point cloud understanding,'' \emph{ECCV}, 2020.

\bibitem{Shen22}
Y.~Shen, L.~Hui, H.~Jiang, J.~Xie, and J.~Yang, ``Reliable inlier evaluation for unsupervised point cloud registration,'' \emph{AAAI}, 2022.

\bibitem{lorax_Elbaz17}
G.~Elbaz, T.~Avraham, and A.~Fischer, ``3d point cloud registration for localization using a deep neural network auto-encoder,'' \emph{CVPR}, 2017.

\bibitem{pointnet_Qi2016}
C.~R. Qi, H.~Su, K.~Mo, and L.~J. Guibas, ``Pointnet: Deep learning on point sets for 3d classification and segmentation,'' \emph{CVPR}, 2017.

\bibitem{fibosampling_Alexa22}
M.~Alexa, ``Super-fibonacci spirals: Fast, low-discrepancy sampling of so(3),'' \emph{CVPR}, 2022.

\bibitem{adam_kingma14}
D.~P. Kingma and J.~Ba, ``Adam: A method for stochastic optimization,'' \emph{ICLR}, 2014.

\bibitem{adamW_Loshchilov19}
I.~Loshchilov and F.~Hutter, ``Decoupled weight decay regularization,'' \emph{ICLR}, 2019.

\bibitem{pytorch_paszke19}
A.~Paszke, S.~Gross, F.~Massa, A.~Lerer, J.~Bradbury, G.~Chanan, T.~Killeen, Z.~Lin, N.~Gimelshein, L.~Antiga, A.~Desmaison, A.~Kopf, E.~Yang, Z.~DeVito, M.~Raison, A.~Tejani, S.~Chilamkurthy, B.~Steiner, L.~Fang, J.~Bai, and S.~Chintala, ``Pytorch: An imperative style, high-performance deep learning library,'' \emph{NeurIPS}, 2019.

\bibitem{manifolds_Lee12}
J.~M. Lee, \emph{Introduction to Smooth Manifolds}.\hskip 1em plus 0.5em minus 0.4em\relax Springer New York, NY, 2012.

\bibitem{topodiff_Hirsh76}
M.~W. Hirsch, \emph{Differential Topology}.\hskip 1em plus 0.5em minus 0.4em\relax Springer New York, NY, 1976.

\bibitem{FeyPyG2019}
M.~Fey and J.~E. Lenssen, ``Fast graph representation learning with {PyTorch Geometric},'' \emph{ICLR}, 2019.

\bibitem{shapenet_Song15}
Z.~Wu, S.~Song, A.~Khosla, F.~Yu, L.~Zhang, X.~Tang, and J.~Xiao, ``3d shapenets: A deep representation for volumetric shapes,'' \emph{CVPR}, 2015.

\bibitem{rpmnet_Yew20}
Z.~J. Yew and G.~H. Lee, ``Rpm-net: Robust point matching using learned features,'' \emph{CVPR}, 2020.

\bibitem{dcp_Wang19}
Y.~Wang and J.~M. Solomon, ``Deep closest point : Learning representations for point cloud registration,'' \emph{ICCV}, 2019.

\bibitem{mac_zhang23}
X.~Zhang, J.~Yang, S.~Zhang, and Y.~Zhang, ``3d registration with maximal cliques,'' \emph{CVPR}, 2023.

\bibitem{mpls_Shi20}
Y.~Shi and G.~Lerman, ``Message passing least squares framework and its application to rotation synchronization,'' \emph{ICML}, 2020.

\bibitem{Bojanic24}
D.~Bojanić, K.~Bartol, J.~Forest, T.~Petković, and T.~Pribanić, ``Addressing the generalization of 3d registration methods with a featureless baseline and an unbiased benchmark,'' \emph{Machine Vision and Applications}, 2024.

\bibitem{dstorm08}
M.~Heilemann, S.~V.~D. Linde, M.~SchuÅNttpelz, R.~Kasper, B.~Seefeldt, A.~Mukherjee, P.~Tinnefeld, and M.~Sauer, ``Subdiffraction-resolution fluorescence imaging with conventional fluorescent probes,'' \emph{Angewandte Chemie International Edition}, 2008.

\bibitem{Wang_yoho22}
H.~Wang, Y.~Liu, Z.~Dong, and W.~Wang, ``You only hypothesize once: Point cloud registration with rotation-equivariant descriptors,'' \emph{ACM MM}, 2022.

\bibitem{Arandjelovic_netvlad18}
R.~Arandjelovic, P.~Gronat, A.~Torii, T.~Pajdla, and J.~Sivic, ``Netvlad: Cnn architecture for weakly supervised place recognition,'' \emph{IEEE TPAMI}, 2018.

\end{thebibliography}

\newpage
\vfill

\end{document}